\documentclass[10pt,journal,compsoc]{IEEEtran}
\usepackage{amsmath,amsfonts}
\usepackage{algorithmic}
\usepackage{algorithm}
\usepackage{array}
\usepackage{textcomp}
\usepackage{stfloats}
\usepackage{url}
\usepackage{verbatim}
\usepackage{graphicx}
\usepackage{cite}
\usepackage{ragged2e}
\usepackage{color}
\usepackage{booktabs}
\usepackage{multirow}
\usepackage{hyperref}
\usepackage{amssymb}
\usepackage{amsmath}
\usepackage{bbding}

\begin{document}

\title{AccDiffusion v2: Towards More Accurate Higher-Resolution Diffusion Extrapolation}

\author{
Zhihang Lin,
Mingbao Lin,
Wengyi Zhan,
Rongrong Ji,~\IEEEmembership{Senior Member,~IEEE}

\IEEEcompsocitemizethanks{

\IEEEcompsocthanksitem Z. Lin and W. Zhan are with the Key Laboratory of Multimedia Trusted Perception and Efficient Computing, Ministry of Education of China, Xiamen University, China. Z.Lin is also with Shanghai Innovation Institute, Shanghai, China. (email:{zhihanglin,zhanwy}@stu.xmu.edu.cn)\protect 
\IEEEcompsocthanksitem M. Lin is with the Rakuten Asia Pte. Ltd., Singapore 048946. (e-mail: linmb001@outlook.com).\protect
\IEEEcompsocthanksitem R. Ji (Corresponding  Author) is with the Key Laboratory of Multimedia Trusted Perception and Efficient Computing, Ministry of Education of China, Xiamen University, China, also with Institute of Artificial Intelligence, Xiamen University, Xiamen 361005, China. (e-mail: rrji@xmu.edu.cn).\protect
}

}





\IEEEtitleabstractindextext{
\begin{abstract}
\justifying
Diffusion models suffer severe object repetition and local distortion when the inference resolution differs from its pre-trained resolution.
We propose AccDiffusion v2, an accurate method for patch-wise higher-resolution diffusion extrapolation without training.
Our in-depth analysis in this paper shows that using an identical text prompt for different patches leads to repetitive generation, while the absence of a prompt undermines image details.
In response, our AccDiffusion v2 novelly decouples the vanilla image-content-aware prompt into a set of patch-content-aware prompts, each of which serves as a more precise description of a patch.
Further analysis reveals that local distortion arises from inaccurate descriptions in prompts about the local structure of higher-resolution images.
To address this issue, AccDiffusion v2, for the first time, introduces an auxiliary local structural information through ControlNet during higher-resolution diffusion extrapolation aiming to mitigate the local distortions.
Finally, our analysis indicates that global semantic information is conducive to suppressing both repetitive generation and local distortion.
Hence, our AccDiffusion v2 further proposes dilated sampling with window interaction for better global semantic information during higher-resolution diffusion extrapolation. 
We conduct extensive experiments, including both quantitative and qualitative comparisons, to demonstrate the efficacy of our AccDiffusion v2.
The quantitative comparison shows that AccDiffusion v2 achieves state-of-the-art performance in image generation extrapolation without training.
The qualitative comparison intuitively illustrates that AccDiffusion v2 effectively suppresses the issues of repetitive generation and local distortion in image generation extrapolation.
Our code is available at \url{https://github.com/lzhxmu/AccDiffusion_v2}.
\end{abstract}

\begin{IEEEkeywords}
Image Generation, High Resolution, Diffusion Model
\end{IEEEkeywords}}

\maketitle
\section{Introduction}\label{sec:intro}

\IEEEPARstart{T}{he} emergence of diffusion models has significantly advanced the generation field, thanks to techniques such as DDPM~\cite{ho2020ddpm}, DDIM~\cite{song2020ddim}, ADM~\cite{dhariwal2021ldm}, and LDMs~\cite{rombach2022SD}. 
These models are known for their outstanding generative abilities and diverse applications.
However, these models perform well only at their pre-trained resolution. 
To generate higher-resolution images, we must train the model at that resolution.
Nonetheless, stable diffusion (SD) models demand extensive high-quality datasets for training and entail tremendous training costs.
For example, SD 1.5 trained with $512\times512$ resolution entails 150,000 A100 GPUs hours~\cite{stable-diffusion-1.5}, while SD 2 trained  with $768\times768$ resolution entails 200,000 A100 GPUs hours~\cite{stable-diffusion-2-1}. 
The training cost is even higher for SDXL~\cite{podell2023sdxl} which is trained with $1024\times1024$ resolution.
The extremely high training cost restricts current open-source SD models to a maximum training resolution of $1024\times1024$~\cite{podell2023sdxl}. 
However, higher-resolution generation finds numerous applications in advertising, gaming, and wallpaper design.
On one hand, large high-resolution image datasets are scarce.
On the other hand, the training cost increases quadratically with resolution.
The above two factors make it infeasible and unaffordable to train ultra-high resolution generative models, such as 4K, directly.
Therefore, exploring how to use pre-trained SD with relatively low resolution for generating ultra-high-resolution images is a valuable research topic for both industry and academia.

\begin{figure}[!t]
    \centering
    \includegraphics[width=\linewidth]{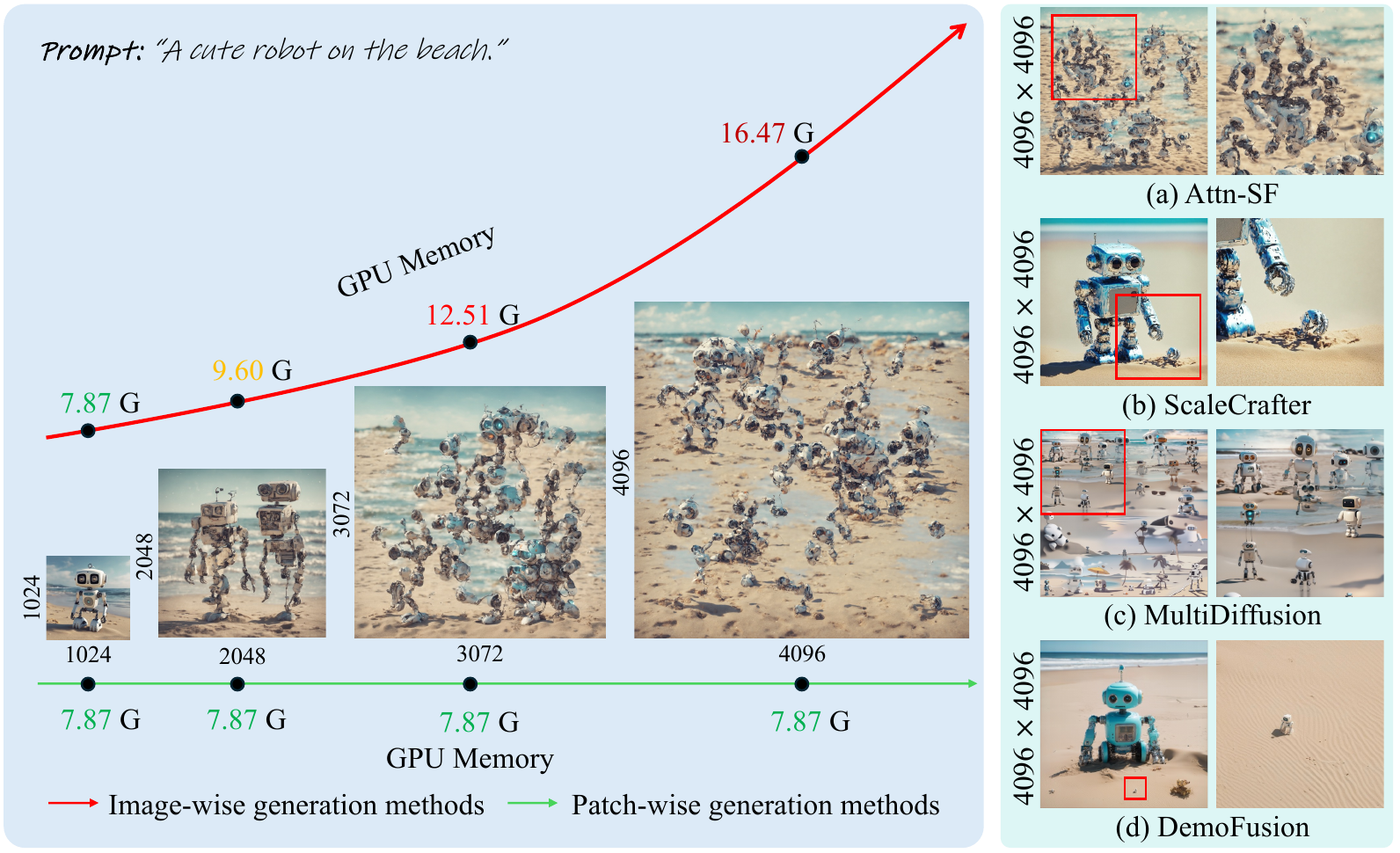}
    \caption{
    Comparison of GPU memory and qualitative results for existing higher-resolution generation methods. The GPU memory of image-wise generation methods, \emph{e.g.}, Attn-SF~\cite{jin2023logn} and ScaleCrafter~\cite{he2023scalecrafter}  greatly increases with resolution. Patch-wise generation methods, \emph{e.g.}, MultiDiffusion~\cite{bar2023multidiffusion} and DemoFusion~\cite{du2023demofusion}  generate images at any resolution with a low GPU memory. Red boxes to highlight the object repetition issue. 
    }
    \label{fig:Related work Comparison}
    \vspace{-2em}
\end{figure}

Recently, there has been an explosive increase in research on image generation extrapolation, using either fine-tuning~\cite{zheng2023any-size-diffusion,xie2023diff-fit} or training-free approaches~\cite{bar2023multidiffusion,du2023demofusion,he2023scalecrafter,lee2023syncdiffusion,jin2023logn,hwang2024upsampleguidance,guo2024make_a_cheap_scaling,zhang2023hidiffusion,kim2024diffusehigh,kim2024beyondscene,lin2024cutdiffusion,tragakis2024pixelsmith,haji2024elasticdiffusion}. 
Previous methods explore image generation extrapolation from various perspectives: attention entropy~\cite{jin2023logn}, frequency-domain~\cite{kim2024diffusehigh}, feature map size~\cite{zhang2023hidiffusion}, and the receptive field of U-Net~\cite{he2023scalecrafter}. 
However, these methods have shown practical limitations in two folds, as illustrated in Fig.\,\ref{fig:Related work Comparison}: (1) GPU memory consumption rises significantly with resolution~\cite{zheng2023any-size-diffusion} and (2) poor image quality~\cite{du2023demofusion}.
Given SD’s ability to generate fine local details, recent works~\cite{du2023demofusion,lee2023syncdiffusion,bar2023multidiffusion,haji2024elasticdiffusion,kim2024beyondscene,tragakis2024pixelsmith,lin2024cutdiffusion} have adopted patch-wise generation to reduce GPU memory usage.
MultiDiffusion~\cite{bar2023multidiffusion} and SyncDiffusion~\cite{lee2023syncdiffusion} merge multiple overlapping patch-wise denoising results to create seamless high-resolution panoramic images. However, applying these techniques to generate higher-resolution, object-focused images often results in repetitive and distorted outputs lacking global semantic coherence, as shown in Fig.\,\ref{fig:Related work Comparison}(c).
ElasticDiffusion~\cite{haji2024elasticdiffusion} uses patch-wise denoising for local signals and incorporates global signals to correct structural distortion, but only supports up to $4\times$ higher resolution. 
DemoFusion~\cite{du2023demofusion} enhances patch-wise image generation extrapolation with global semantic information through residual connections and dilated sampling.
Despite partially addressing repetitive object generation, it still suffers from small object repetition and local distortion in ultra-high-resolution images, as shown in Fig.\,\ref{fig:Related work Comparison}(d).
In summary, off-the-shelf patch-wise denoising methods fail to accurately extrapolate generation to higher resolutions compared to the pre-trained resolutions, mainly resulting from two issues: (1) repetitive generation~\cite{bar2023multidiffusion,lee2023syncdiffusion,du2023demofusion} and (2) local distortion~\cite{du2023demofusion}. 
Therefore, how to accurately generate a higher-resolution image in a patch-wise manner remains an unresolved challenge.

\begin{figure}[tb]
    \centering
    \includegraphics[width=\linewidth]
    {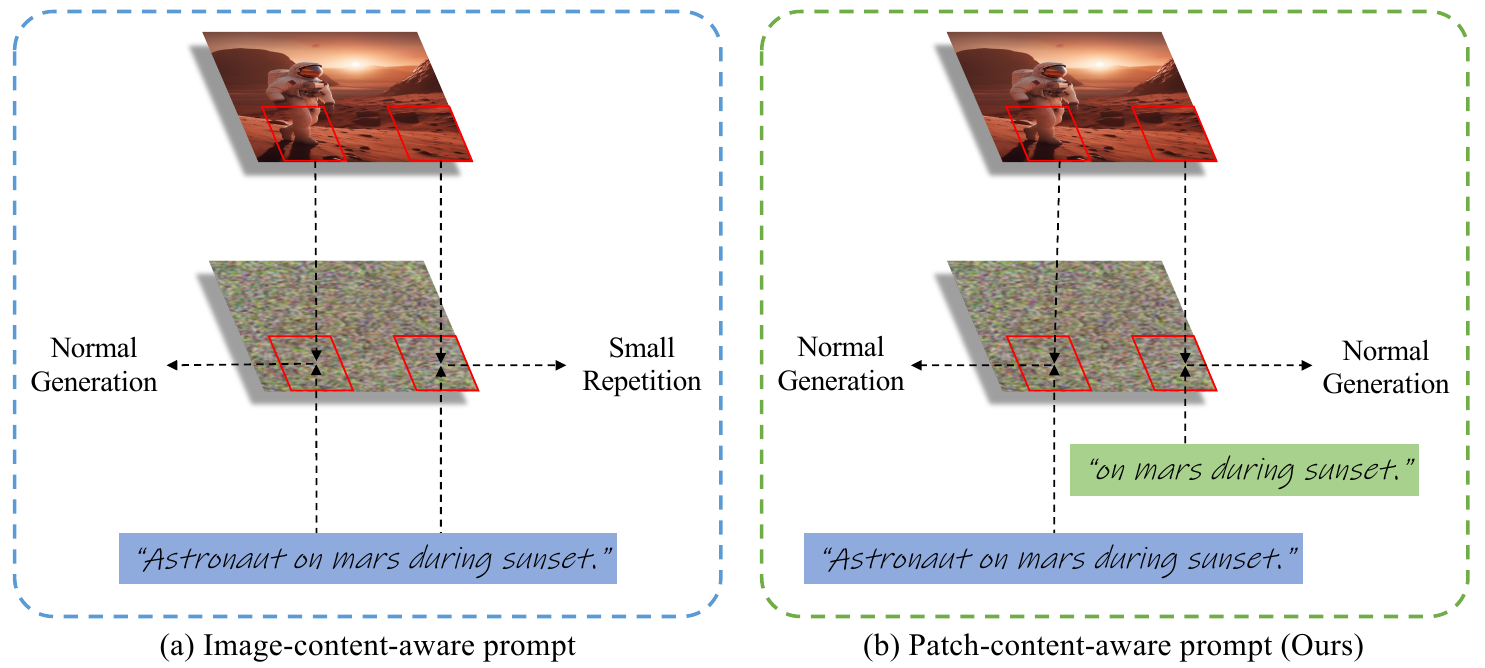}
    \caption{Image-content-aware prompt \emph{v.s.} Patch-content-aware prompt.}
    \label{fig:repetition analyze}
    \vspace{-1em}
\end{figure}

In this paper, we propose AccDiffusion v2 to conduct more accurate higher-resolution diffusion extrapolation, effectively suppressing issues of repetitive generation and local distortion.
First, our in-depth analysis indicates, as illustrated in Fig.\,\ref{fig:repetition analyze}(a), small object repetitive generation is the adversarial outcome of an identical text prompt on all patches, encouraging to generate repetitive objects, and global semantic information, suppressing the generation of repetitive objects.
Hence, we propose to decouple the vanilla image-content-aware prompt into a set of patch-content-aware substrings, each of which serves as a more precise prompt to describe the patch contents. Specifically, we utilize the cross-attention map from the low-resolution generation process to determine whether a word token should serve as the prompt for a patch.
If a word token has a high response in the cross-attention map region corresponding to the patch, it should be included in the prompt, and vice versa.
Secondly, we find that patch-content-aware prompt suppresses the repetitive generation effectively, but local distortion persists in higher-resolution images. 
We further analyze that local distortion is the adversarial outcome of inaccurate prompts, encouraging to generate overall structures, and global semantic information, encouraging to generate local structures.
Hence, we provide an additional structure condition for patch-wise generation to suppress the influence of inaccurate prompts.
Specifically, we inject the structure information of low-resolution generation into stable diffusion during patch-wise denoising through ControlNet~\cite{zhang2023controlnet}, suppressing distortion well. 
Finally, recent works~\cite{du2023demofusion,haji2024elasticdiffusion} show that accurate global semantic information is conducive to suppressing repetitive generation and local distortion simultaneously.
Previous work~\cite{du2023demofusion} uses dilated sampling to provide global semantic information for higher-resolution generation.
However, we observe that the conventional dilated sampling generates globally inconsistent and noisy information, disrupting the generation of higher-resolution images. 
Such inconsistency stems from the independent denoising of dilation samples without interaction. 
In response, we employ a position-wise bijection function to enable interaction between the noise from different dilation samples. 
Experiments show that our dilated sampling with interaction leads to smoother global semantic information, as shown in Fig.\,\ref{fig:ablation analyze}(d). 

We conduct both extensive qualitative and quantitative experiments to confirm the efficacy of AccDiffusion v2. The qualitative results show its success in suppressing repetitive generation and local distortion in higher-resolution image generation.
Quantitative results also highlight its top performance in training-free image generation extrapolation. Additionally, we perform comprehensive ablation studies to assess the individual contributions of the three modules proposed in AccDiffusion v2, validating their role in enhancing overall performance.

The conference version of this paper, termed AccDiffusion, is published in~\cite{lin2024accdiffusion}. 
In the conference version, our contributions are summarized as follows:
(1) We identify the reason for repetitive generation during patch-wise denoising and introduce patch-content-aware prompts to effectively suppress this issue.
(2) We propose dilated sampling with interaction to generate more accurate global semantic information, effectively reducing repetitive generation.

In this journal version, we build upon the conference version and introduce several substantial improvements as follows:
(1) We conduct a more in-depth analysis of the causes of local distortion during patch-wise denoising. To address this issue, we incorporate low-resolution structural information into the patch-wise denoising process using ControlNet, effectively mitigating local distortions and addressing the limitations raised in the conference version.
(2) We perform more extensive experiments, both quantitatively and qualitatively. Quantitatively, we compare AccDiffusion v2 with the latest related methods, achieving updated state-of-the-art performance. Qualitatively, we provide comparisons with more related methods (Fig.\,\ref{fig:Qualitative Comparison}) and showcase additional results at various resolutions (Fig.\,\ref{fig:more_visualization}). Furthermore, we reconduct ablation studies (Sec.\,\ref{sec:ablation on core modules}) to validate the compatibility of our new module with those proposed in the conference version. We also apply AccDiffusion v2 to different stable diffusion variants (Fig.\,\ref{fig:variants}).
(3) While this study provides valuable insights, it is not without its shortcomings. Therefore, we include an analysis of failure cases (Fig.\,\ref{fig:failure cases}), identify the limitations of this work, and suggest potential directions for future research.

\section{Related Work}

\subsection{Diffusion Models}
Probabilistic generative models like DDPM~\cite{ho2020ddpm}, DDIM~\cite{song2020ddim}, and LDMs~\cite{rombach2022SD} are diffusion models that transform Gaussian noise into samples through iterative denoising. DDPM stands out for its impressive image generation ability, leveraging Markovian forward and reverse processes. DDIM further enhances DDPM by employing non-Markovian reverse processes, cutting down sampling time significantly. 
By integrating the diffusion process into latent space, LDMs achieve more efficient training and inference. 
Consequently, several open-source LDMs-based stable diffusion models have achieved state-of-the-art performance in image synthesis.
This progress has led to widespread applications of diffusion models across various generative tasks, including image~\cite{dhariwal2021ldm,ho2020ddpm,nichol2021improved,song2020ddim,saharia2022photorealistic}, audio~\cite{huang2023make-a-audio,ghosal2023Text-to-Audio}, video~\cite{singer2022make-a-video,ho2022imagen}, and 3D object~\cite{lin2023magic3d,xu2023dream3d,poole20223d-dreamfusion}, \emph{etc}.

\subsection{Training-Free Higher-Resolution Image Generation} 

While stable diffusion delivers remarkable results,   the high training cost limits it to low resolutions, leading to low-quality images when the inference resolution differs from the training resolution~\cite{jin2023logn,he2023scalecrafter,du2023demofusion}. Recent studies explore using pre-trained diffusion models to generate higher-resolution images. These approaches are two folds: image-wise generation~\cite{he2023scalecrafter,jin2023logn,hwang2024upsampleguidance,guo2024make_a_cheap_scaling,zhang2023hidiffusion,kim2024diffusehigh} and patch-wise generation~\cite{bar2023multidiffusion,lee2023syncdiffusion,du2023demofusion,haji2024elasticdiffusion,lin2024cutdiffusion,kim2024beyondscene,tragakis2024pixelsmith}.

Image-wise generation methods either directly~\cite{he2023scalecrafter,jin2023logn,hwang2024upsampleguidance,zhang2023hidiffusion} or gradually~\cite{guo2024make_a_cheap_scaling,kim2024diffusehigh} scale the input of diffusion models to the target resolution before applying forward and reverse processes on the latent space.
These methods often require architectural modifications, such as adjusting the attention scale factor~\cite{jin2023logn}, the feature map size of U-Net~\cite{zhang2023hidiffusion}, and the receptive field of convolutional kernels~\cite{he2023scalecrafter}, to prevent repetitive generation. SelfCascade~\cite{guo2024make_a_cheap_scaling} and DiffuseHigh~\cite{kim2024diffusehigh} upsample the generated pre-trained resolution images and refine their details through forward and reverse processes. UG~\cite{hwang2024upsampleguidance} employs a pre-trained diffusion model with an additional term called upsample guidance during sampling to create higher-resolution images.
However, these methods often fail to achieve the desired high-resolution details and encounter out-of-memory errors when generating ultra-high resolution images (\emph{e.g.}, 8K) on consumer-grade GPUs due to the exponential increase in memory requirements as the latent space size increases.

Patch-wise generation produces higher-resolution images through patch-wise denoising and can generate images of any resolution on consumer-grade GPUs.
However, these methods~\cite{bar2023multidiffusion,lee2023syncdiffusion} struggle with object repetition and local distortion. 
Du \emph{et al.}\cite{du2023demofusion} and Tragakis \emph{et al.}~\cite{tragakis2024pixelsmith} attempt to reduce repetitive generation by incorporating global structural information from lower-resolution images.
Haji-Ali \emph{et al.}\cite{haji2024elasticdiffusion} separate high-resolution image generation into local and global signals to address distortion but only support up to 4$\times$ higher resolution.
Lin \emph{et al.}\cite{lin2024cutdiffusion} split the patch-wise denoising process into comprehensive structure denoising and specific detail refinement to tackle the local repetition issue. 
Kim \emph{et al.}\cite{kim2024beyondscene} use a staged and hierarchical approach for human-centric scenes.

\section{Backgrounds}\label{sec:backgrounds}

\textbf{Latent Diffusion Models (LDMs)}.
LDMs perform the diffusion process in latent space.
For an image $\mathbf{x}_0 \in \mathbb{R}^{H\times W \times 3}$, an autoencoder $\mathcal{E}$ encodes it into latent space as:
\begin{equation}
    \mathbf{z}_0 = \mathcal{E}(\mathbf{x}_0),
\end{equation}
where $\mathbf{z}_0 \in \mathbb{R}^{h\times w \times c}$ is the latent representation of an image. Then the diffusion process of LDMs can be formulated as:
\begin{equation} \label{eq:diffusoin}
    \mathbf{z}_t = \sqrt{\bar{\alpha}_t}\mathbf{z}_0 + \sqrt{1-\bar{\alpha}_t}\varepsilon,\quad \varepsilon\sim \mathcal{N}(0,\mathbf{I}),
\end{equation}
where $\{{\alpha}_t\}_{t=1}^T$  is a set of prescribed variance schedules and $\bar{\alpha}_t = \Pi_{i=1}^t \alpha_i$. Then a network $\varepsilon_{\theta}$ is trained to perform conditional sequential denoising by predicting added noise, with the training objective defined as follows:
\begin{equation}
\label{SD_training_loss}
    \underset{\theta}{\text{min}}\ \mathbb{E}_{\mathbf{z}_0, \varepsilon \sim \mathcal{N}(0,1), t}\Big[\left\| \varepsilon - \varepsilon_{\theta}\big(\mathbf{z}_t, t, \tau_{\theta}(y)\big)\right\|_2^2\Big],
\end{equation}
in which $t \sim \text{Uniform}(1, T)$, $\tau_{\theta}(y)\in \mathbb{R}^{M\times d_{\tau}}$ is an intermediate representation of condition $y$ and $M$ is the number of word tokens in the prompt $y$.
In the cross-attention of U-Net, $\tau_{\theta}(y)$ is subsequently mapped to keys and values as:
\begin{equation}
\label{cross-attention}
\begin{aligned}
        Q = W_Q\cdot \varphi(z_t),\quad K = W_K\cdot \tau_{\theta}(y),\quad V = W_V\cdot \tau_{\theta}(y), \\  
    \mathcal{M} = \text{Softmax}(\frac{QK^T}{\sqrt{d}}),\quad \text{Attention}(Q,K,V) = \mathcal{M} \cdot V.
\end{aligned}
\end{equation}
Here $\varphi(z_t)\in \mathbb{R}^{N\times d_\epsilon}$ represents an intermediate noise representation within the U-Net.
And $N = h \times w$ denotes the pixel number of the latent noise $z_t$.
The matrices $W_Q \in \mathbb{R}^{d\times d_{\epsilon}}, W_K \in \mathbb{R}^{d\times d_{\tau}} $, and $W_V\in \mathbb{R}^{d\times d_{\tau}}$ are learnable projections, while $\mathcal{M} \in \mathbb{R}^{N\times M}$ is the cross-attention maps. 
Without loss of generality, we omit the expression of multi-head cross-attention for conciseness.

During denoising process, diffusion model estimates the noise in  $\mathbf{z}_{t}$ and recovers the cleaner version $\mathbf{z}_{t-1}$  through:
\begin{equation}
\label{denoising process}
\begin{split}
    \mathbf{z}_{t-1} &= \hat{\alpha} \cdot \mathbf{z}_t + \hat{\beta} \cdot \varepsilon_{\theta}\big(\mathbf{z}_t,t,\tau_{\theta}(y)\big), \\
    \hat{\alpha} &= \sqrt{\frac{\alpha_{t-1}}{\alpha_t}},\\
    \hat{\beta} &= \Bigg(\sqrt{\frac{1}{\alpha_{t-1}} - 1} - \sqrt{\frac{1}{\alpha_t} - 1} \Bigg).
\end{split}
\end{equation}

By iteratively denoising through Eq.\,(\ref{denoising process}), a noise-free latent $\mathbf{z}_0$ is decoded to image $\mathbf{x}_0$ through decoder $\mathcal{D} ( \cdot )$ as:
\begin{equation}
    \mathbf{x}_0 = \mathcal{D}(\mathbf{z}_0).
\end{equation}

\textbf{ControlNet}.
For controllable image generation, ControlNet~\cite{zhang2023controlnet} adds an additional condition encoder on pre-trained diffusion models. 
The denoising process of ControlNet can be represented as follows:
\begin{equation}
\label{denoising process controlnet}
\begin{split}
    \mathbf{z}_{t-1} &= \hat{\alpha}\cdot\mathbf{z}_t + \hat{\beta}\cdot \varepsilon_{\theta'}\big(\mathbf{z}_t,t,\tau_{\theta}(y),q\big).\\
\end{split}
\end{equation}

Here, $\varepsilon_{\theta'}$ represents ControlNet and $q$ denotes extra conditions, such as canny edges~\cite{canny1986canny}, human poses~\cite{humanpose}, depth maps~\cite{depthmap}. Note that the introduced ControlNet is a plug-and-play extension without altering the parameters of pre-trained diffusion models.

\textbf{Patch-wise Denoising}.
MultiDiffusion~\cite{bar2023multidiffusion} first uses a shift window to sample overlapped patches and then fuses the denoising results to generate higher-resolution images.
The sampling progress can be formulated as:
\begin{equation}
\label{eq:sample_patch}
    \{\mathbf{z}_t^i\}_{i=1}^{P_1} = \text{Sample}(\mathcal{Z}_t, d_h, d_w),
\end{equation}
where $\mathcal{Z}_t \in \mathbb{R}^{h'\times w'\times c}$ is the latent representation of a higher-resolution image.
$\mathbf{z}_t^i \in \mathbb{R}^{h\times w\times c}$ denotes the sampled patches, where $h' > h$ and $w' > w$, and the total patch count $P_1 = (\frac{h'-h}{d_h}+1)\times(\frac{w'-w}{d_w}+1)$. $d_h$ and $d_w$ represent vertical and horizontal strides, respectively.
Subsequently, the cleaner version $\{\mathbf{z}_{t-1}^i\}_{i=1}^{P_1}$ is obtained by denoising $\{\mathbf{z}_{t-1}^i\}_{i=1}^{P_1}$ using Eq.\,(\ref{denoising process}).  
Finally, MultiDiffusion fuses patches $\{\mathbf{z}_{t-1}^i\}_{i=1}^{P_1}$ to get $\mathcal{Z}_{t-1}$, where the overlapped parts take the average.
A higher-resolution image is then obtained by directly decoding $\mathcal{Z}_0$ into the image $\mathbf{X}_0$.
Building upon MultiDiffusion, DemoFusion~\cite{du2023demofusion} further includes a progressive upscaling strategy to incrementally generate higher-resolution images, residual connections to maintain global consistency with the lower-resolution image by injecting an intermediate noise-inversed representation, and dilated sampling to enhance the global semantic information of higher-resolution images.

\begin{figure*}[!t]
    \centering
    \includegraphics[width=\linewidth]{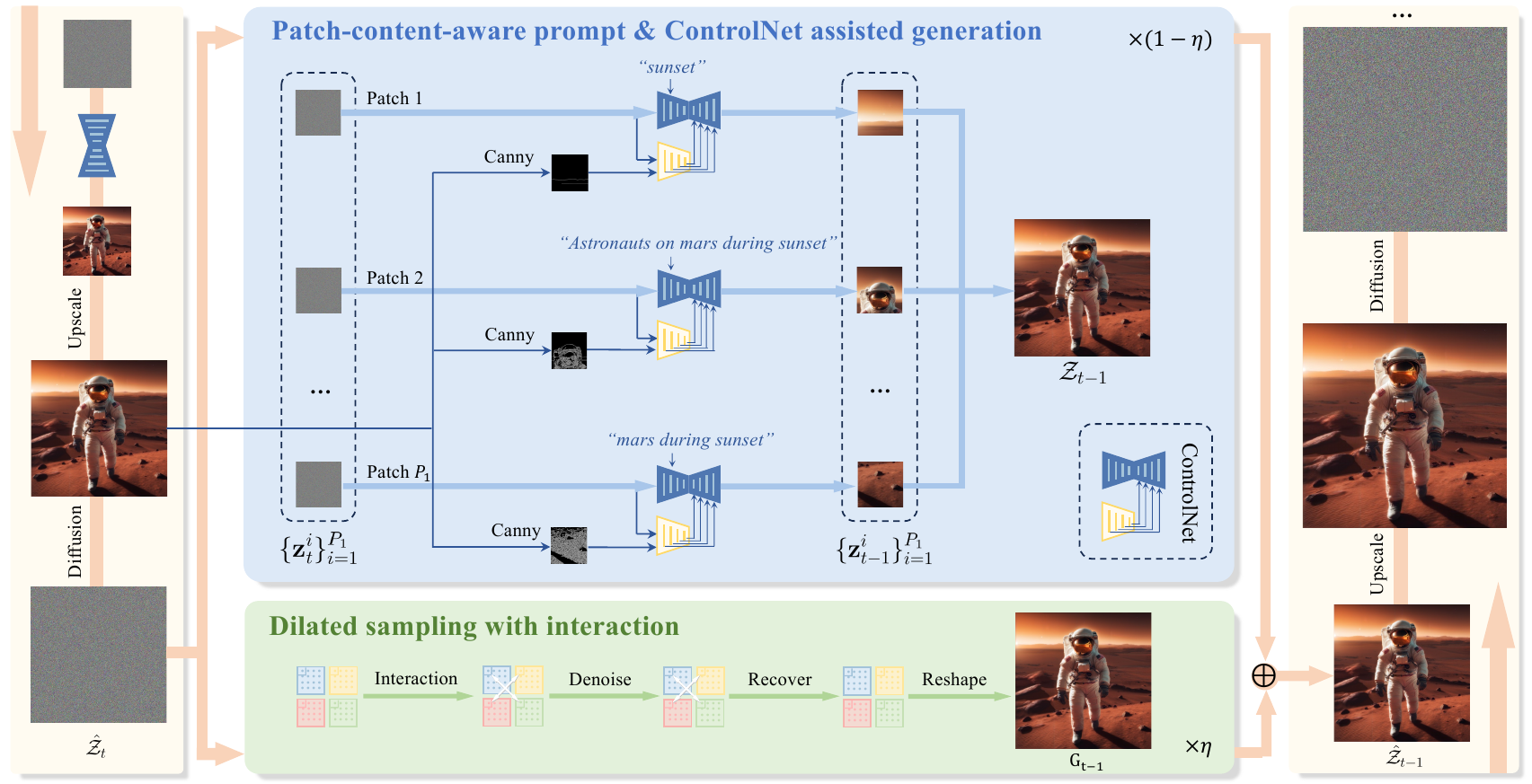}
    \caption{
    The framework of AccDiffusion v2 simplified by setting the denoising step $T = 1$ for illustration. All operations are operated within the latent space. Firstly, the pre-trained diffusion model conducts a full denoising progress at the pre-trained resolution to obtain the denoised latent. This latent is then upscaled to a higher resolution and undergoes diffusion progress as per Eq.\,(\ref{eq:diffusoin}). During higher-resolution image generation, AccDiffusion v2 utilizes patch-content-aware prompts, ControlNet-assisted generation, and dilated sampling with interaction to suppress repetitive generation and local distortion until the target resolution is reached.}
    \label{fig:framework}
\end{figure*}

\section{AccDiffusion v2}
This section formally introduces AccDiffusion v2, a plug-and-play extension for diffusion models that enables accurate higher-resolution image generation.
Similar to recent works~\cite{bar2023multidiffusion,du2023demofusion,kim2024diffusehigh}, AccDiffusion v2 adapts a progressive recipe to conduct image generation extrapolation in a patch-wise fashion, which can generate ultra-high resolution images on one consumer-grade GPU.
The framework of AccDiffusion v2 is illustrated in Fig.\,\ref{fig:framework}.
The major differences between AccDiffusion v2 and recent methods are three folds:
(1) AccDiffusion v2 uses patch-content-aware prompts for each patch to conduct accurate higher-resolution image generation, while recent works~\cite{du2023demofusion,kim2024diffusehigh} use image-content-aware prompt for all patches.
(2) AccDiffusion v2 innovatively integrates ControlNet~\cite{zhang2023controlnet} during the patch-wise denoising to alleviate local distortion.
(3) AccDiffusion v2 uses dilated sampling with interaction to generate accurate global semantic information, while recent methods~\cite{du2023demofusion} independently denoise dilation samples without interaction.

\subsection{Patch-Content-Aware Prompts}\label{sec:patch-content-aware prompts}

\begin{figure}[!t]
    \centering
    \includegraphics[width=\linewidth]{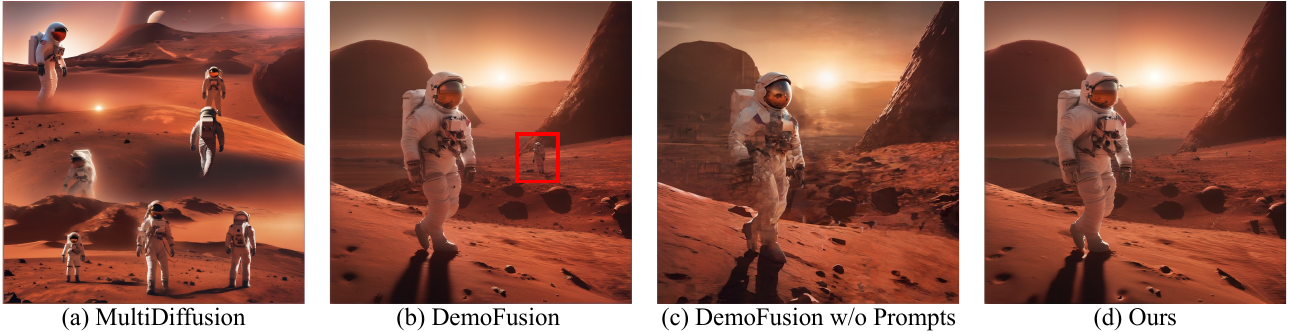}
    \caption{Results of higher-resolution image generation. 
    (a) MultiDiffusion suffers from large object repetition.
    (b) DemoFusion suffers from small object repetition.
    (c) DemoFusion without text prompts results in a severe loss of details.
    (d) Our method, utilizing patch-content-aware prompts, effectively suppresses small object repetition while preserving image fidelity.
    Best viewed by zooming in.    \label{fig:ablation analyze}
    }
    \vspace{-1em}
\end{figure}

Current image generation extrapolation methods~\cite{bar2023multidiffusion,du2023demofusion} suffer from repetitive generation as shown in Fig.\,\ref{fig:Related work Comparison} and Fig.\,\ref{fig:ablation analyze}. 
This repetition typically falls into two categories: large object repetition and small object repetition.
For example, MultiDiffusion~\cite{bar2023multidiffusion}, which use the same prompts across all patches, generates multiple large repeated objects scattered across the image. 
By integrating residual connection \& dilated sampling, DemoFusion~\cite{du2023demofusion} addresses large object repetition but leads to small object repetition in the background as shown in Fig.\,\ref{fig:ablation analyze}(b).
To identify the cause of small object repetition, we perform an ablation experiment by excluding the text prompt during higher-resolution generation of DemoFusion. 
The result in  Fig.\,\ref{fig:ablation analyze}(c) shows that the removal of prompts completely eliminates small repetitive objects but results in a noticeable loss of detail.
From these results, it is reasonable to conclude that small object repetition arises as an adverse effect from using the same text prompt across all patches, as well as from residual connection and dilated sampling operations.
While the former promotes object repetition (Fig.\,\ref{fig:ablation analyze}(a)), the latter diminishes it (Fig.\,\ref{fig:ablation analyze}(b)).
As a result, DemoFusion tends to generate small repetitive objects.
The above analysis reveals that simply excluding text prompts during higher-resolution generation to eliminate small object repetition is not a feasible remedy, as it would inevitably result in a compromise on image fidelity.
Considering the significant role that text prompts play in image generation and the inaccuracy of identical text prompts for all patches, it is crucial to tailor more accurate prompts for each patch.
That is, if an object is not present in a patch, the corresponding word in the text prompts should not serve as a prompt for that patch.

Bearing the above conclusion in mind, we explore patch-content-aware substring set $\{ \gamma^i\}_{i=1}^{P_1}$ of the entire text prompt, each of which injects a condition into their respective patches.
It is challenging to know in advance what content a patch generates.
Luckily, recent works~\cite{du2023demofusion,kim2024diffusehigh,tragakis2024pixelsmith} leverage residual connections to incorporate global information from low-resolution images into high-resolution image generation, resulting in a higher-resolution image that retains a similar structure to the low-resolution one. 
This inspires us to infer patch content directly from the low-resolution image.
One direct but cumbersome method is to manually examine the patch content within the low-resolution image and then define a prompt for each patch, which undermines the usability of diffusion models.
Alternatively, SAM~\cite{kirillov2023sam} could be applied to segment the upscaled low-resolution image and verify object presence within each patch, but this introduces significant storage and computational demands.
How to generate patch-content-aware prompts without external models is the key to success.

\begin{figure}[!t]
    \centering
    \includegraphics[width=\linewidth]{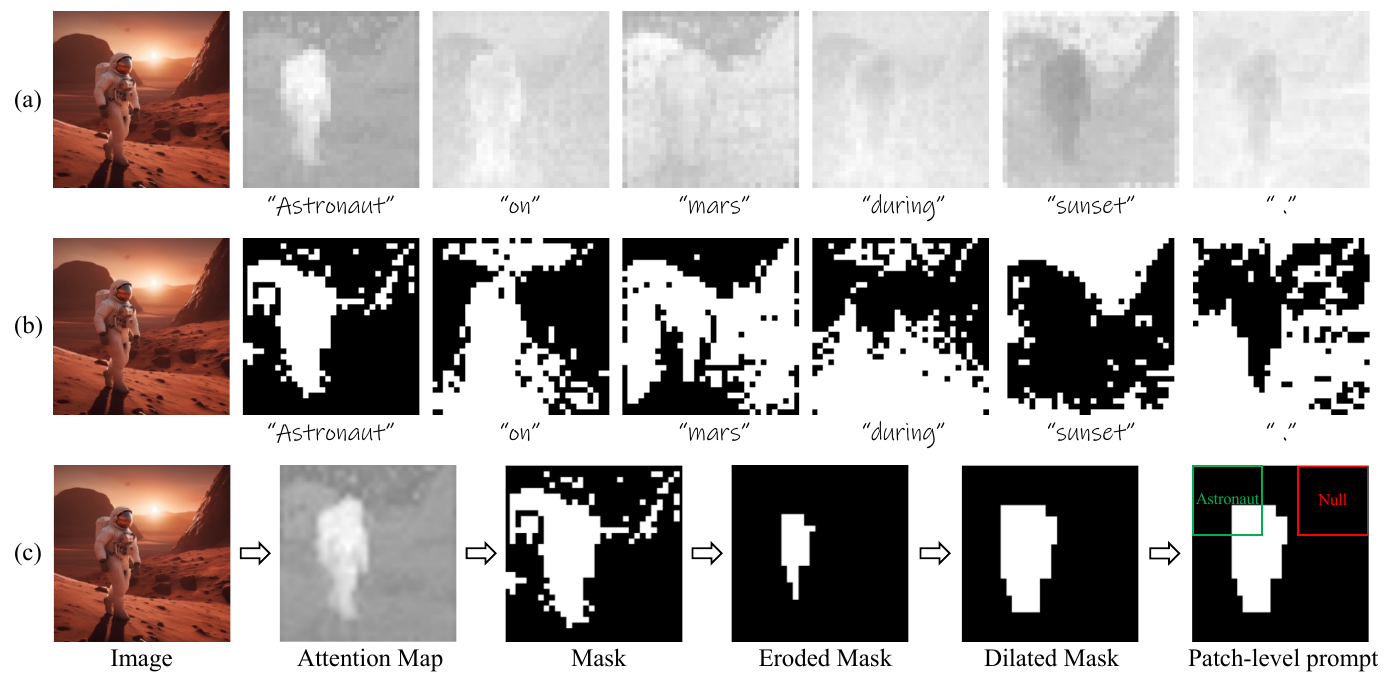}
    \caption{
    Visualization of averaged attention map from the up blocks and down blocks in U-Net. We reshape the attention map into a 2D shape before visualization.
    (a) Cross-attention map visualization using open source code~\cite{hertz2022prompt-to-prompt}.  
    (b) Highly responsive regions of each word. 
    (c) The illustration of the patch-level prompt generation process, including morphological operations to eliminate small connected areas. Here we use the word ``Astronaut'' as an example. All words in the prompt will go through the above process. Best viewed by zooming in.}
    \label{fig:attention_map}
    \vspace{-1em}
\end{figure}

Drawing inspiration from image editing~\cite{hertz2022prompt-to-prompt}, we shift our focus to the cross-attention maps in low-resolution generation $\mathcal{M} \in \mathbb{R}^{N \times M}$, to derive patch-content-aware prompts.
Here, $N$ is the pixel number of the latent noise $z_t$ and $M$ is the number of word tokens in the prompt $y$.
The column $\mathcal{M}_{:, j}$ indicates how much the latent noise attends to the $j$-th word token.
The basic principle is simple: the attentiveness ($\mathcal{M}_{i, j}$) of image regions is mostly higher than others if it is attended by the $j$-th word token, as shown in Fig.\,\ref{fig:attention_map}(a). 
To identify the highly relevant region of each word token, we convert the attention map $\mathcal{M}$ into a binary mask $\mathcal{B} \in \mathbb{R}^{N \times M}$ as follows:
\begin{equation}\label{eq:highly responsive regions}
\mathcal{B}_{i,j} = \left\{
\begin{array}{ll}
1 , \textrm{\; if  $\mathcal{M}_{i,j}$ $>$ $\overline{\mathcal{M}}_{:,j}$},\\
0 , \textrm{\; otherwise},
\end{array}\right.
\end{equation}
where $i$ and $j$ enumerate $N$ and $M$, respectively. 
The threshold $\overline{\mathcal{M}}_{:,j}$ is the mean of $\mathcal{M}_{:,j}$, as discussed in Sec.\,\ref{ablaiton study}.
Regions with values above this threshold are classified as highly responsive, while those below are less responsive.
Next, we reshape word-level masks $\{\mathcal{B}_j\}^M_{j=1}$ as follows:
\begin{equation}
\hat{\mathcal{B}}_j = \text{Reshape}(\mathcal{B}_{:,j},(h_a,w_a)), 
\end{equation}
where $h_a = \frac{h}{s}$ and $w_a = \frac{w}{s}$ denote the height and width of the attention map, respectively. $h$ and $w$ are the height and width of the noise.
The factor ``$s$'' represents the down-sampling scale in the corresponding U-Net model block.
The mask ${\mathcal{B}}_j$  for the $j$-th word token is reshaped into a 2D shape for subsequent operations.

However, as shown in Fig.\,\ref{fig:attention_map}(b), many small connected areas appear in highly responsive regions ${\mathcal{B}}_j$.
To reduce the impact of these small connected areas, we use the opening operation $\mathcal{O}(\cdot)$ from mathematical morphology~\cite{soille1999morphological}, resulting in the final mask for each word, as shown in Fig.\,\ref{fig:attention_map}(c).
The resulting processed masks $\{\Tilde{\mathcal{B}}_j\}^M_{j=1}$ are defined as:
\begin{equation}
 \Tilde{\mathcal{B}}_j = \mathcal{O}(\hat{\mathcal{B}}_j) = \omega(\delta(\hat{\mathcal{B}}_j)),
\end{equation}
where $\delta(\cdot)$ and $\omega(\cdot)$ denote the erosion and dilation operations, respectively.
We then interpolate $ \Tilde{\mathcal{B}}_j \in \mathbb{R}^{h_a\times w_a}$ to $ \Tilde{\mathcal{B}}'_j \in \mathbb{R}^{h'_a\times w'_a}$, where $h'_a = \frac{h'}{s}$ and $w'_a = \frac{w'}{s}$.
Recall that $h'$ and $w'$ are the sizes of higher-resolution latent representation as defined in Sec.\,\ref{sec:backgrounds}.
Similar to Eq.\,(\ref{eq:sample_patch}), we use a shifted window to sample patches from $\Tilde{\mathcal{B}}'_j$ , resulting in a series of patch masks $\{\{\mathbf{m}^i_j\}_{i=1}^{P_1}\}_{j=1}^{M}$, where $\mathbf{m}^i_j \in \mathbb{R}^{h_a\times w_a}$ and $P_1$ is the total number of patches. 
It is important to note that each $\mathbf{m}^j_i$ corresponds to a specific patch noise $\mathbf{z}_t^i$.

Recall that if an object is not present in a patch, the corresponding word token in the text prompts should not serve as a prompt for that patch.
With this in mind, we can determine the patch-content-aware prompt $\gamma^i$, a sub-sequence of prompt $y$, for each patch $\mathbf{z}_t^i$ as follow:

\begin{equation}
\label{eq:patch-content-aware prompt}
\left\{
\begin{array}{ll}
y_j \in \gamma^i, \textrm{\; if $\frac{\sum(\mathbf{m}^i_j)_{:,:}}{h_a \times w_a}$ $>$ $c$},\\
y_j \notin \gamma^i, \textrm{\; otherwise},
\end{array}\right.
\end{equation}
where $j$ and $i$ enumerates $M$ and $P_1$, respectively. 
The hyper-parameter $c\in (0,1)$ determines if the proportion of a highly responsive region corresponding to a word $y_j$ exceeds the threshold for inclusion in the prompts of patch $z_t^i$.
We then concatenate all words that should appear in a patch together, resulting in patch-content-aware prompts $\{\gamma^i\}_{i=1}^{P_1}$ for noise patches $\{z_t^i\}_{i=1}^{P_1}$ during patch-wise denoising.
Patch-content-aware prompts effectively suppress small object repetition while preserving image fidelity (Fig.\,\ref{fig:ablation analyze}(d)).

\subsection{More Accurate Generation of Local Content}
\label{sec:controlnet}
\begin{figure}[!t]
    \centering
    \includegraphics[width=\linewidth]
    {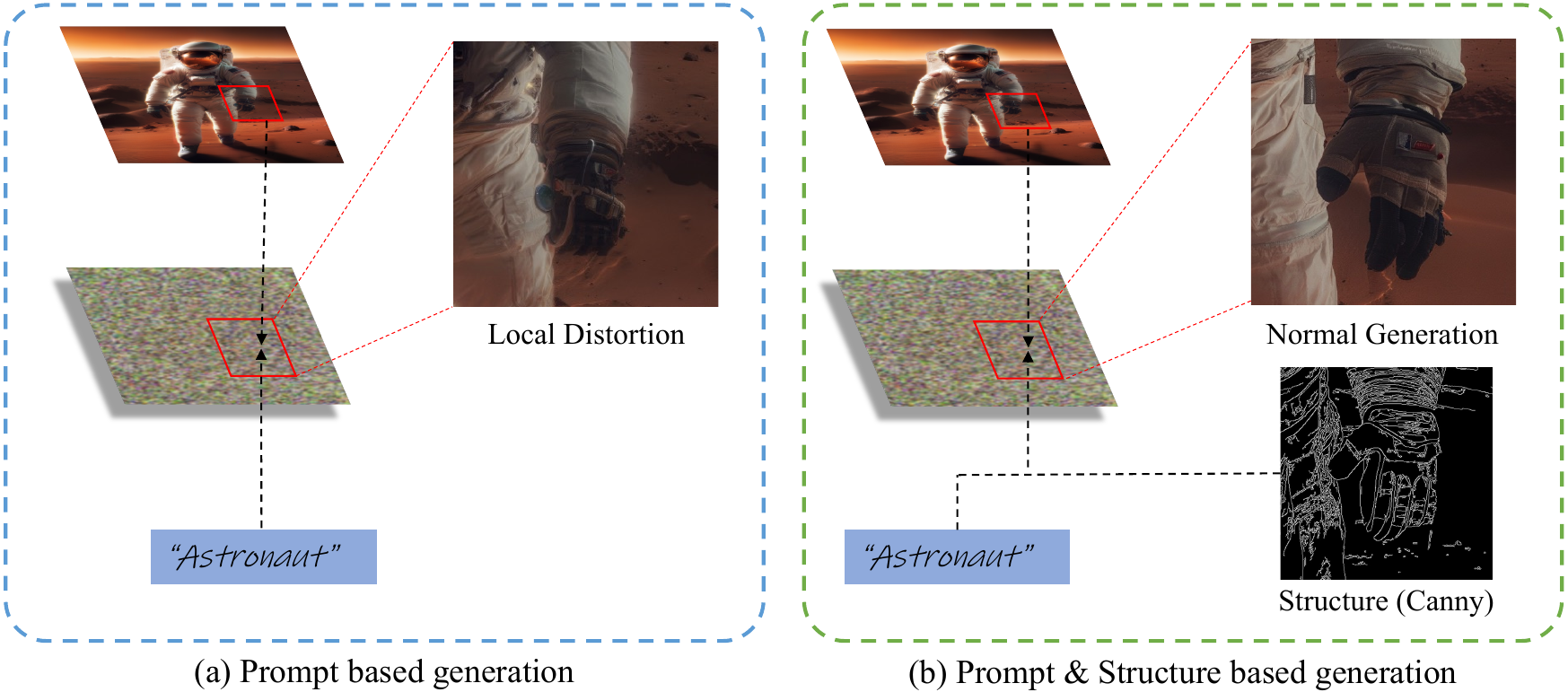}
    \caption{
    Prompt based generation \emph{v.s.} Prompt \& Structure based generation.
    Best viewed by zooming in.}
    \label{fig:local irrational analyze}
\end{figure}

Patch-content-aware prompts effectively suppress the repetitive generation in higher-resolution diffusion extrapolation~\cite{lin2024accdiffusion}.
Despite this improvement, local distortion persists in the results, as illustrated in Fig.\,\ref{fig:local irrational analyze}(a).
Drawing parallels to the analysis in Sec\,\ref{sec:patch-content-aware prompts}, we speculate that the patch-content-aware prompts are not enough to accurately describe the content of the patches.
In Fig.\,\ref{fig:local irrational analyze}(a), we use the patch corresponding to the astronaut's hand to give an in-depth analysis. 
This patch tends to generate a complete structure (astronaut) conditioned by the word ``astronaut'' in the prompt, but global semantic information tends to generate local structures (hand).
Consequently, the clash between the two leads to a local distortion.
A simplistic remedy would involve excluding the inaccurate prompt during higher-resolution diffusion extrapolation.
However, we have demonstrated in Sec.\ref{sec:patch-content-aware prompts} that prompts significantly contribute to the details of results, playing a crucial role in image generation.
Therefore, the challenge of local distortion must be approached from another perspective while retaining the prompt.

As the structure of images in pre-trained resolution is rational, the structure of relatively low-resolution images can serve as a reference during higher-resolution diffusion extrapolation.
First, the denoised latent $z_0 \in \mathbb{R}^{h\times w \times c}$ is decoded to low-resolution image $I = \mathcal{D}(z_0) \in \mathbb{R}^{H\times W \times 3}$.  
Next, the image $I$ is interpolated to higher resolution $I' \in \mathbb{R}^{H'\times W' \times 3}$ with $H'>H $ and $W' > W$.
Subsequently, the canny edge detector~\cite{canny1986canny} is used to detect the edges $C \in \mathbb{R}^{H'\times W' \times 3}$ in images $I'$.
Similar to Eq.\,(\ref{eq:sample_patch}), we use a shifted window to sample patches from $C$, resulting in a series of patches $\{\mathcal{C}^i\}_{i=1}^{P_1}$, where $\mathcal{C}^i \in \mathbb{R}^{H\times W \times 3}$ and $P_1$ is the total number of patches.
So far, each patch $\mathbf{z}^i_t$ has a corresponding prompt $\gamma^i$ and local structure information $\mathcal{C}^i$.
By integrating the ControlNet~\cite{zhang2023controlnet} $\varepsilon_{\theta'}$, the denoising process of patch $\mathbf{z}^i_t$ can be expressed as:
\begin{equation}
\label{denoising process accdiffusonv2}
\begin{split}
    \mathbf{z}_{t-1} &= \hat{\alpha}\cdot \mathbf{z}_t + \hat{\beta} \cdot \varepsilon_{\theta'}\big(\mathbf{z}_t,t,\tau_{\theta'}(\gamma^i), \mathcal{C}^i \big).\\
\end{split}
\end{equation}

We enable high-fidelity generation of higher-resolution images by incorporating ControlNet, benefiting both the details from the patch-content-aware prompts $\gamma^i$ and precise local structures from the local structure information $\mathcal{C}^i$.

\begin{figure}[!t]
    \centering
    \includegraphics[width=\linewidth]{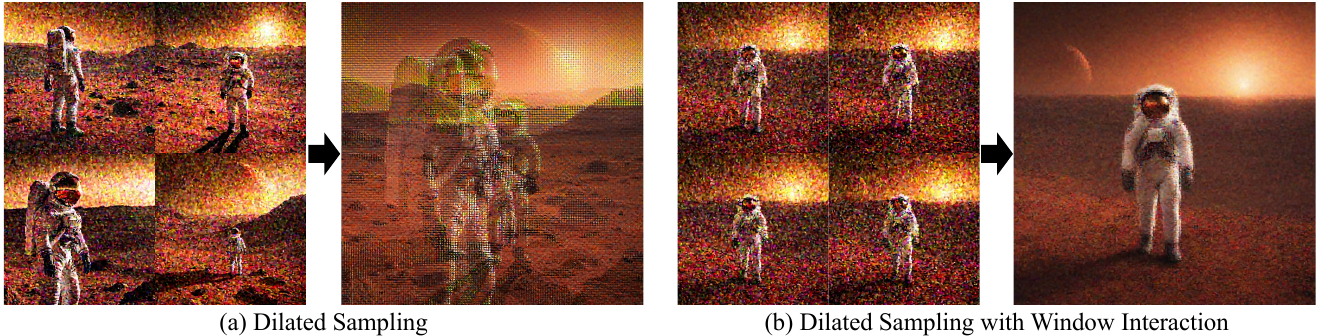}
    \caption{
        Results of $4\times$ higher-resolution image generation. (a) The result of dilated sampling without window interaction. (b)The result of our dilated sampling with window interaction. Best viewed by zooming in.}    \label{fig:ablation_window_interaction}

    \vspace{-1em}
\end{figure}

\begin{figure}[!t]
    \centering
    \includegraphics[width=\linewidth]{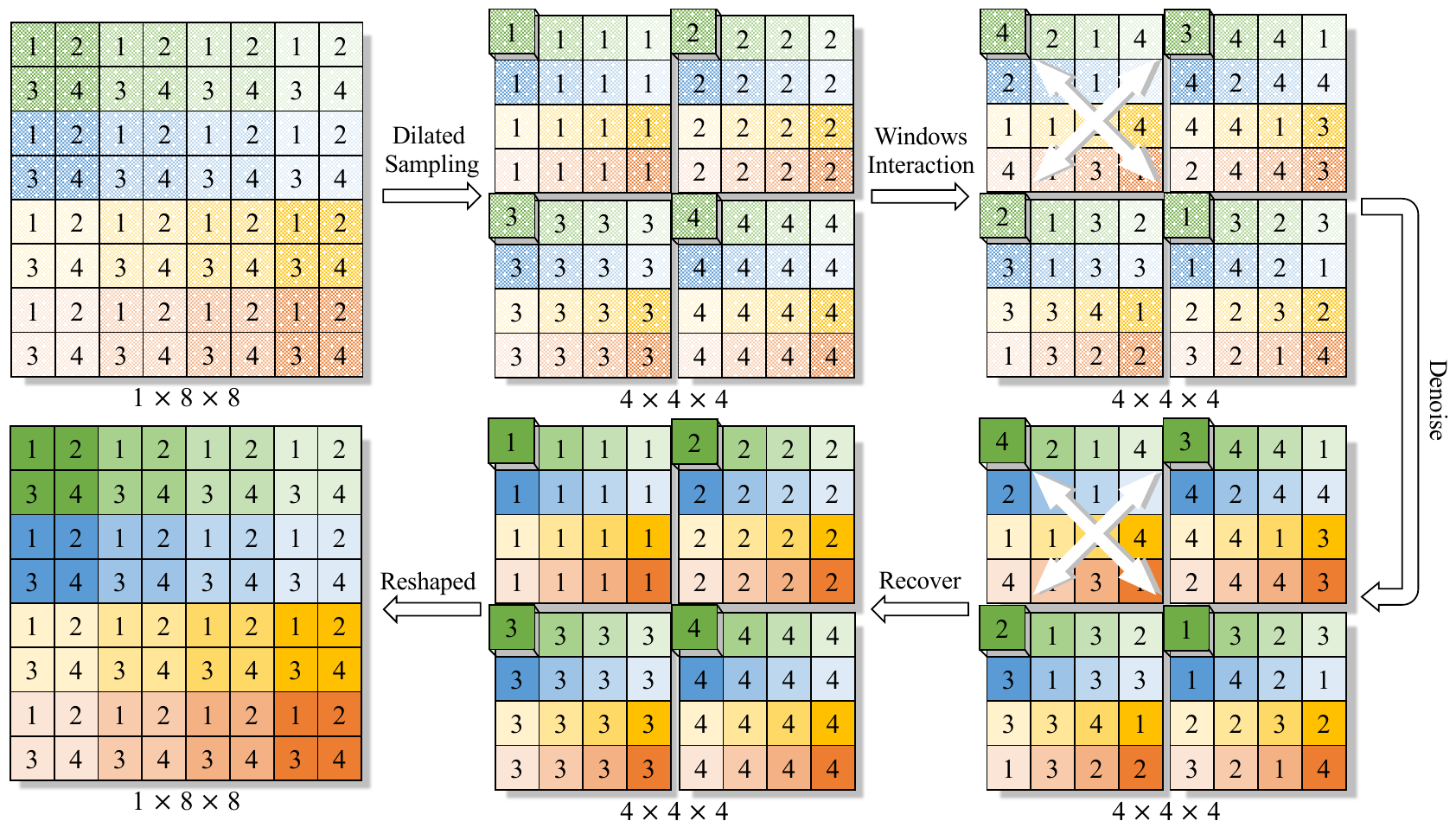}
    \caption{Illustration of dilated sampling with window interaction: $8 \times 8$ higher-resolution and $4 \times 4$ low-resolution. 
    The numbers $\{1,2,3,4\}$ represent the different positions within the same window (same color). The interaction operation is conducted in the window.}
    \label{fig:dilated sampling pipline}
\end{figure}

\subsection{ Dilated Sampling with  Window Interaction}
\label{Shuffle Dilated-Sampling}
Both our analysis in Sec.\,\ref{sec:patch-content-aware prompts} and recent works~\cite{du2023demofusion,haji2024elasticdiffusion} show that global semantic information effectively suppresses object repetition.
Dilated sampling is a feasible way to inject global semantic information during higher-resolution extrapolation~\cite{du2023demofusion}.
Given a higher-resolution latent representation $\mathcal{Z}_t \in \mathbb{R}^{h'\times w'\times c}$, a set of patch samples $\{D_t^k\}_{k=1}^{P_2}$ are dilated sampled as:
\begin{equation}
    \mathcal{D}_t^{k} = (\mathcal{Z}_{t})_{i::h_s,j::w_s,:},
\end{equation}
where $k$ is defined as $k = i \times w_s + j + 1$, ranging from $1$ to $P_2$. 
The indices $i$ and $j$ vary from $0$ to $h_s-1$ and $w_s-1$, respectively.
The sampling stride is calculated as $h_s = \frac{h'}{h}$ and $w_s= \frac{w'}{w}$, with $\{h',w'\}$ and $\{h, w\}$ representing the height and width of higher and low resolution latent representation.
DemoFusion performs denoising on $\mathcal{D}_t$ independently via Eq.\,(\ref{denoising process}) to obtain $\mathcal{D}_{t-1} \in \mathbb{R}^{P_2 \times  h \times w \times c}$.
Next, the denoised outputs $\{\mathcal{D}_{t-1}^k\}_{k=1}^{P_2}$ are combined to reconstruct $G_{t-1} \in \mathbb{R}^{h'\times w' \times c}$, which are added to patch-wise denoised latent representation ${\mathcal{Z}_{t-1}}$ as:
\begin{equation}
\label{dilated sampling }
\begin{aligned}
    {\hat{\mathcal{Z}}_{t-1}} = (1 - \eta) &\cdot {\mathcal{Z}_{t-1}} + \eta \cdot {G_{t-1}}, 
\end{aligned}
\end{equation}
where $(G_{t-1})_{i::h_s,j::w_s,:}  = {\mathcal{D}^{k}_{t-1}}$ and $\eta$ decreases from $1$ to $0$ following a cosine schedule.
As shown in Fig.\,\ref{fig:ablation_window_interaction}(a), the global semantic information generated by dilated sampling appears inconsistent and non-smooth. This issue arises because dilated sampling independently denoises $\{D_t^k\}_{k=1}^{P_2}$ and subsequently combines the denoised outputs $\{\mathcal{D}_{t-1}^k\}_{k=1}^{P_2}$ to form the global semantic information. The lack of interaction among different samples $\mathcal{D}_t^{k}$, as illustrated in Fig.\,\ref{fig:ablation_window_interaction}(a), results in discrepancies between the denoised outputs, leading to non-smooth global semantic information.
To solve this issue, as illustrated in Fig.\,\ref{fig:dilated sampling pipline}, we enable window interaction among different samples  prior to each denoising process through a bijective function:
\begin{equation}
\begin{split}
{\mathcal{D}_t}^{k,h,w} &= {\mathcal{D}_t}^{f_t^{h,w}(k),h,w}, \\
f^{h,w}_t:\{1,2,\cdots,P_2\} &\Rightarrow \{1,2,\cdots,P_2\},
\end{split}
\end{equation}
where $f^{h,w}_t$ is a bijective function with random mapping, with the mapping varying on the specific position or time step.
We then perform standard denoising progress on $ \{\mathcal{D}_{t}^k\}_{k=1}^{P_2}$  to obtain $\{\mathcal{D}_{t-1}^k\}_{k=1}^{P_2}$.
Before applying Eq.\,(\ref{dilated sampling }) to $\{\mathcal{D}_{t-1}^k\}_{k=1}^{P_2}$, we recover the position by using the inverse mapping ${(f^{h,w}_t)}^{-1}$  of $f^{h,w}_t$  as:
\begin{equation}
\begin{split}
{\mathcal{D}_{t-1}}^{k,h,w} &= {\mathcal{D}_{t-1}}^{{(f^{h,w}_t)}^{-1}(k),h,w}, \\ {(f^{h,w}_t)}^{-1}:\{1,2,\cdots,P_2\} &\Rightarrow \{1,2,\cdots,P_2\},
\end{split}
\end{equation}
which yields more smooth global semantics like Fig.\,\ref{fig:ablation_window_interaction}(b).
This verifies that through window interaction, different samples $\{D_t^k\}_{k=1}^{P_2}$ can exchange information with each other, which is crucial for generating more consistent and smoother global semantic information.

\section{Experimentation}

\subsection{Experimental Setup}
Since AccDiffusion v2 has not been fine-tuned on any higher-resolution image dataset, we select only training-free comparison methods, including: 
SDXL-DI~\cite{podell2023sdxl},
Attn-SF~\cite{jin2023logn},
ScaleCrafter~\cite{he2023scalecrafter},
MultiDiffusion~\cite{bar2023multidiffusion},
HiDiffusion~\cite{zhang2023hidiffusion},
DiffuseHigh~\cite{kim2024diffusehigh},
DemoFusion~\cite{du2023demofusion},
and AccDiffusion~\cite{lin2024accdiffusion}.
Although both image super-resolution and diffusion extrapolation aim to generate high-resolution images, they differ in that one uses images as input and the other uses text. 
Thus, we did not compare AccDiffusion v2 with super-resolution methods. 
Previous works have shown that super-resolution generates inferior details than diffusion extrapolation methods~\cite{he2023scalecrafter,du2023demofusion}.
To verify the effectiveness of AccDiffusion v2, we select the widely used SDXL~\cite{podell2023sdxl} for quantitative and qualitative comparisons.
For quantitative comparison, we set the hyperparameter $c$ to 0.3.
The ControlNet checkpoint used is available on Hugging Face at {\href{https://huggingface.co/xinsir/controlnet-canny-sdxl-1.0}{https://huggingface.co/xinsir/controlnet-canny-sdxl-1.0}}.

\subsection{Quantitative Comparison}
\label{quantitative comparison}
We employ three widely-used metrics: Frechet Inception Distance (FID)~\cite{heusel2017fid}, Inception Score (IS)~\cite{salimans2016is}, and CLIP Score~\cite{radford2021clip-score} for quantitative evaluations. 
Specifically, $\text{FID}_r$ assesses the Frechet Inception Distance between generated high-resolution images and real images, while $\text{IS}_r$ calculates the Inception Score for these generated high-resolution images. Notably, both $\text{FID}_r$ and $\text{IS}_r$ require resizing images to $299^2$ resolutions, which may not provide optimal assessments for high-resolution images. To address this, inspired by methods ~\cite{du2023demofusion,chai2022Any-resolution-training}, we crop 10 local patches at native resolution ($1\times$) from each generated high-resolution image before resizing, yielding $\text{FID}_c$ and $\text{IS}_c$.
The CLIP Score is calculated based on the cosine similarity between image embeddings and text prompts, providing an additional alignment metric.
For quantitative comparison, we randomly selected $10,000$ images from the Laion-5B dataset~\cite{schuhmann2022laion-5b} as the real image set and used $1,000$ randomly chosen text prompts from Laion-5B as input for AccDiffusion v2, generating a corresponding set of high-resolution images.

AccDiffusion v2 achieves state-of-the-art performance in diffusion extrapolation tasks, as shown in  Table\,\ref{tab:quantitative comparison}. 
More accurate patch-content-aware prompts, enhanced accuracy in local content generation, and improved integration of global structure information enabled by dilated sampling with interaction contribute to the improvements. 
These improvements are especially effective for high-resolution image generation (16$\times$).
In comparison with other training-free image generation extrapolation methods, AccDiffusion v2 produces quantitative results that more closely align with those at pre-trained resolutions, underscoring its robust extrapolation capabilities in generating high-quality images beyond pre-trained resolutions. The inference time of AccDiffusion v2 is slightly higher than that of AccDiffusion due to the additional cost of suppressing local distortion through ControlNet~\cite{zhang2023controlnet}.
Note that $\text{FID}$, $\text{IS}$, and CLIP-Score may not directly indicate the presence of repetitive generation or local distortion in the generated images. Therefore, we perform a qualitative comparison in next section to confirm the efficacy of AccDiffusion v2 in reducing such artifacts.

\begin{table}[!tb]
  \caption{Comparison of quantitative metrics between different training-free image generation extrapolation methods. We use \textbf{bold} to emphasize the best result and \underline{underline} to emphasize the second best result. 
  }
  \label{tab:quantitative comparison}
    \resizebox{\linewidth}{!}{
  \setlength\tabcolsep{2pt}
  \centering
  \begin{tabular}{@{}clcccccc@{}}
    \toprule
    Resolusion & Method & $\text{FID}_r\downarrow $ & $\text{IS}_r\uparrow$  & $\text{FID}_c\downarrow$ & $\text{IS}_c\uparrow$ & CLIP$\uparrow$ & Time\\
    \midrule
    1024 $\times$ 1024 (1$\times$)    & SDXL-DI & 58.49 & 17.39 & 58.08 & 25.38 & 33.07 & $<$1 min \\
    \midrule
    \multirow{9}{*}{2048 $\times$ 2048 (4$\times$)} & SDXL-DI& 124.40 & 11.05 & 88.33 & 14.64 & 28.11 & 1 min\\
                                                    & Attn-SF & 124.15 & 11.15 & 88.59 & 14.81 & 28.12 & 1 min \\
                                                    & MultiDiffusion & 81.46 & 12.43 & 44.80 & 20.99 & 31.82 & 2 min\\
                                                    & ScaleCrafter & 99.47 & 12.52 & 74.64 & 15.42 & 28.82 & 1 min\\ 
                                                    & HiDiffusion &  87.77           &   14.99          &      59.80             &  21.31               &    28.89       &  1 min \\
                                                    & DiffuseHigh &  62.51           &  16.35           &     40.22              &      21.72           &   32.58        &  1 min \\
                                                    & DemoFusion & 60.46 & 16.45 & 38.55 & 24.17 & 32.21 & 3 min\\ 
                                                    & AccDiffusion & \underline{59.63} & \underline{16.48} & \underline{38.36} & \underline{24.62} & \underline{32.79} & 3 min\\
                                                    & AccDiffusion v2 & \textbf{58.12} & \textbf{18.62} & \textbf{38.10} & \textbf{25.59} & \textbf{32.84} & 4 min\\
                                                    
    \midrule
    \multirow{9}{*}{3072 $\times$ 3072 (9$\times$)}  & SDXL-DI & 170.61 & 7.83 & 112.51 & 12.59 & 24.53 & 3 min\\
                                                    & Attn-SF & 170.62 & 7.93 & 112.46 & 12.52 & 24.56 & 3 min\\
                                                    & MultiDiffusion & 101.11 & 8.83 & 51.95 & 17.74 & 29.49 & 6 min\\
                                                    & ScaleCrafter  & 131.42 & 9.62 & 105.79 & 11.91 & 27.22 & 7 min\\  
                                                    & HiDiffusion & 136.73           & 10.06             & 100.86            & 13.59              & 26.20             & 2 min \\
                                                    & DiffuseHigh &  62.43           &   15.51           &  44.96            &  18.28             &  32.65             & 3 min \\
                                                    & DemoFusion & 62.43 & 16.41 & 47.45 & 20.42 & 32.25 & 11 min\\ 
                                                    & AccDiffusion & \underline{61.40} & \underline{17.02} & \underline{46.46} & \underline{20.77} & \underline{32.82} & 11 min\\
                                                    & AccDiffusion v2 & \textbf{58.78} & \textbf{18.36} & \textbf{44.90} & \textbf{21.05} & \textbf{32.84} & 15 min\\
    \midrule
    \multirow{9}{*}{4096 $\times$ 4096 (16$\times$)} & SDXL-DI & 202.93 & 6.13 & 119.54 & 11.32 & 23.06 & 9 min\\
                                                    & Attn-SF & 203.08 & 6.26 & 119.68 & 11.66 & 23.10 & 9 min\\
                                                    & MultiDiffusion & 131.39 &  6.56 &  61.45 & 13.75 & 26.97 & 10 min\\
                                                    & ScaleCrafter & 139.18 &  9.35 & 116.90 & 9.85 & 26.50 & 20 min\\     
                                                    & HiDiffusion &  145.98         & 8.54               &  172.58           &  7.69             &  24.08            & 3 min \\
                                                    & DiffuseHigh &   64.12         &  14.68             &    57.97          &  15.08            &  \textbf{33.75}            & 8 min \\
                                                    & DemoFusion & 65.97 & 15.67 & 59.94 & 16.60 & 33.21 & 25 min\\ 
                                                    & AccDiffusion & \underline{63.89} & \underline{16.05} & \underline{58.51} & \underline{16.72} & \underline{33.79} & 26 min\\        
                                                    & AccDiffusion v2 & \textbf{60.88} & \textbf{17.21} & \textbf{57.63} & \textbf{16.78} & 32.83 & 35 min\\
  \bottomrule
  \end{tabular}
  }
\end{table}

\begin{figure*}[!t]
    \centering
    \includegraphics[width=0.9\linewidth]{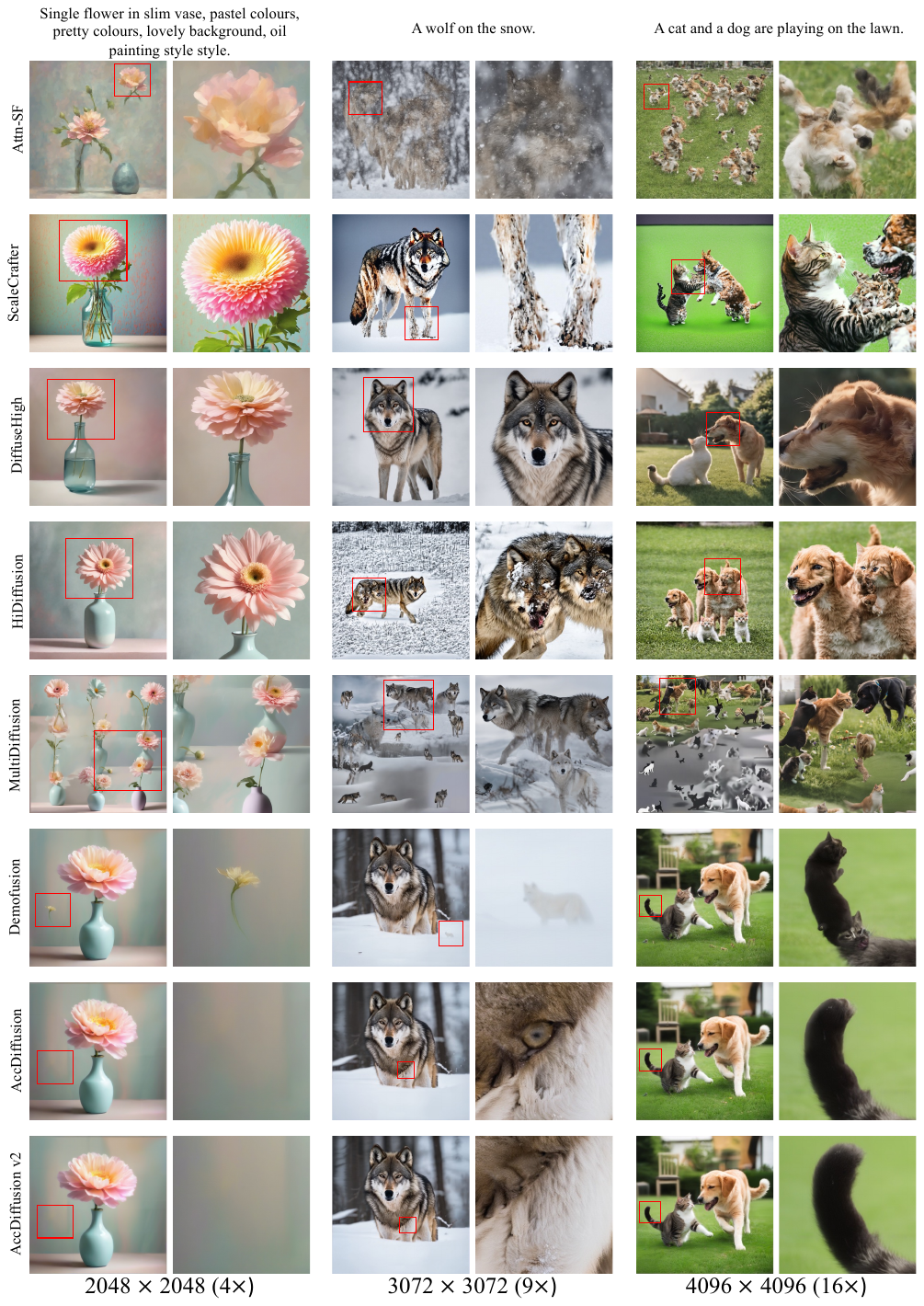}
    \caption{Qualitative comparison of our AccDiffusion with existing training-free image generation extrapolation methods~\cite{jin2023logn,he2023scalecrafter,kim2024diffusehigh,zhang2023hidiffusion,bar2023multidiffusion,du2023demofusion,lin2024accdiffusion}. We upscale the red box region for better observation. Best viewed zoomed in.
    }
    \label{fig:Qualitative Comparison}
\end{figure*}

\begin{figure}[htbp]
    \centering
    \includegraphics[width=1\linewidth]{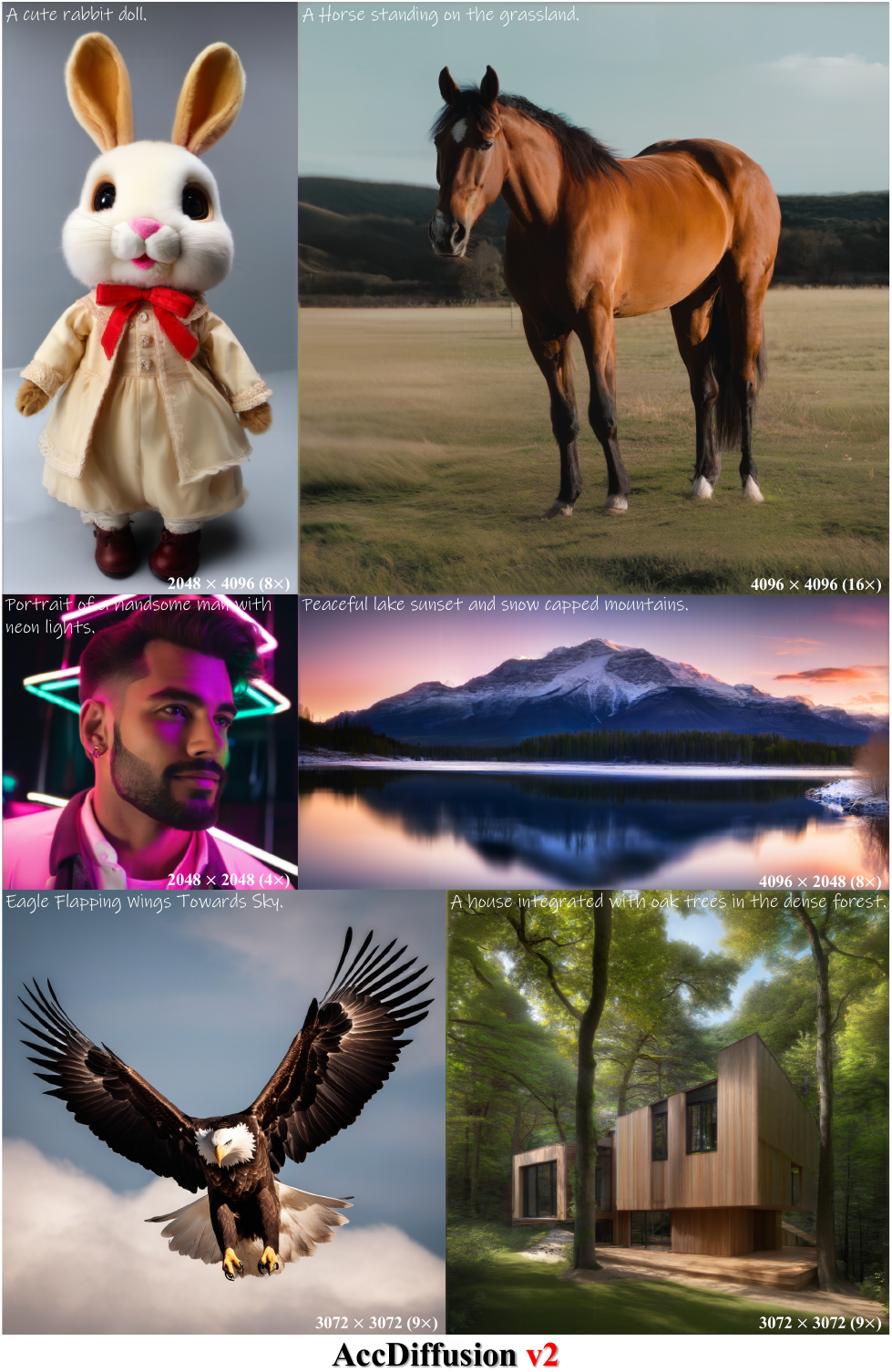}
    \caption{
     More selected results of AccDiffusion v2 at various resolutions. Best viewed by zooming in.
     }
    \label{fig:more_visualization}
\end{figure}
 
\subsection{Qualitative Comparison}
Fig.\,\ref{fig:Qualitative Comparison} shows a comparison between AccDiffusion v2 and other training-free text-to-image generation extrapolation methods, including Attn-sf~\cite{jin2023logn}, ScaleCrafter~\cite{he2023scalecrafter}, DiffuseHigh~\cite{kim2024diffusehigh},
HiDiffusion~\cite{zhang2023hidiffusion},
MultiDiffusion~\cite{bar2023multidiffusion}, 
DemoFusion~\cite{du2023demofusion},
and AccDiffusion~\cite{lin2024accdiffusion}.
As the resolution increases, Attn-SF suffers from severe structural distortion and a significant decline in visual quality.
ScaleCrafter avoids object repetition but experiences detail degradation at $3072 \times 3072$ resolution and structural distortions at $4096 \times 4096$ resolution, as highlighted in the red box.
DiffuseHigh can generate high-fidelity images at $2048 \times 2048$ and $3072 \times 3072$ resolutions, but it still suffers from local distortion at the higher resolution of $4096 \times 4096$, also highlighted in the red box. 
Though HiDiffusion is an efficient image generation
extrapolation method but suffers from severe object repetition and local distortion at high resolutions, such as $3072 \times 3072$ and $4096 \times 4096$. 
MultiDiffusion can generate seamless images but also suffers from significant repetitive and distorted generation.
DemoFusion tends to generate small repetitive objects, like the small wolf at $3072 \times 3072$ and small cats and dogs at $4096 \times 4096$, with the frequency of repetition escalating with image resolution. 
It also suffers local distortion, such as the tail of the cat at $4096 \times 4096$, both of which significantly degrade image quality.
AccDiffusion demonstrates superior performance in generating high-resolution images without such repetitions.
However, it still suffers from local distortion in the foreground, such as the eye on the leg of the wolf and the strange shape of the cat's tail.
In contrast, AccDiffusion v2 can conduct more accurate higher-resolution extrapolation without repetitions or local distortion, leading to high-quality results. 
We provide more results in Fig.\,\ref{fig:more_visualization}, demonstrating that AccDiffusion v2 can produce impressive results across various resolutions, aspect ratios, and subjects.

\begin{figure}[!t]
    \centering
    \includegraphics[width=\linewidth]{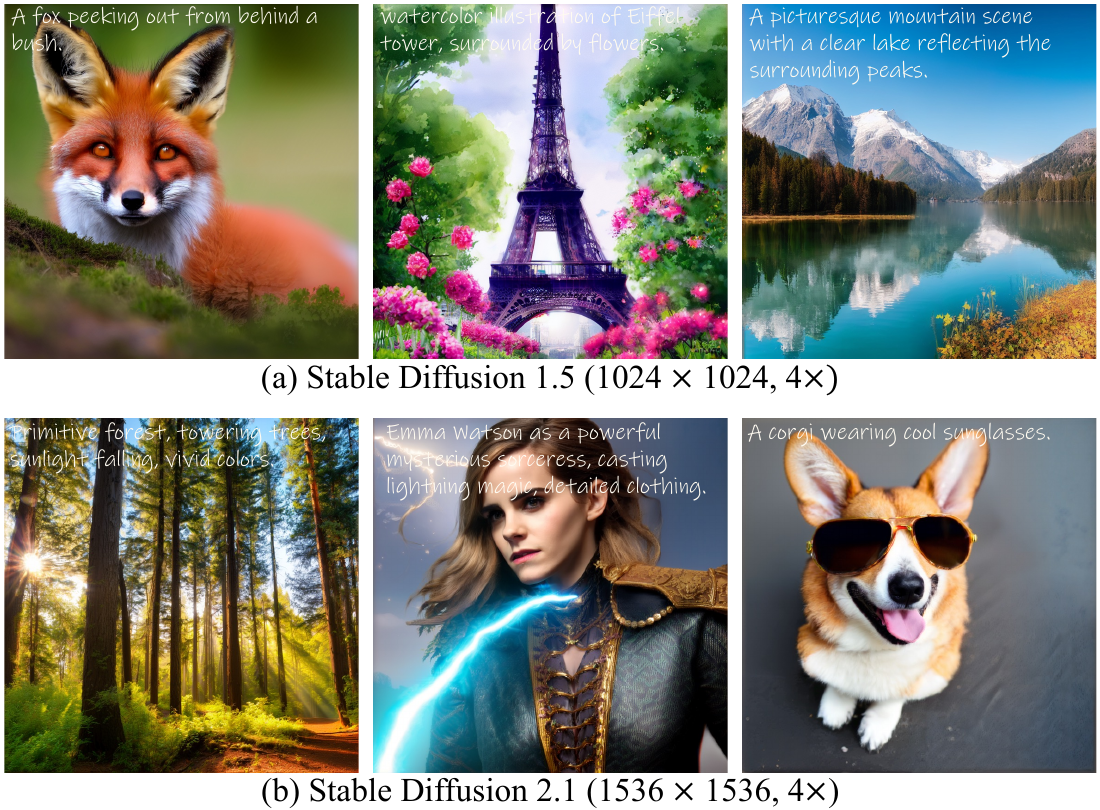}
    \caption{Results of AccDiffusion v2 on other stable diffusion variants: (a) Stable diffusion 1.5 (default resolution of $512^2$) and (b) Stable diffusion 2.1 (default resolution of $768^2$). All images are generated at $4 \times$ resolution. Best viewed by zooming in.
    }
    \label{fig:variants}
\end{figure}

\subsection{More Stable Diffusion Variants}
AccDiffusion v2 is a plug-and-play framework that can be easily used to conduct higher-resolution diffusion extrapolation for different diffusion models. 
Thus, we implement AccDiffusion v2 for other latent diffusion models (LDMs), specifically Stable Diffusion 1.5 (SD 1.5)\cite{stable-diffusion-1.5} and Stable Diffusion 2.1\cite{stable-diffusion-2-1} (SD 2.1).
As demonstrated in Fig.\,\ref{fig:variants}, AccDiffusion v2 effectively generates high-resolution images without noticeable repetition or localized distortion. 
However, it’s crucial to consider that AccDiffusion v2’s results are influenced by the foundational quality of the LDMs used.
Consequently, the visual fidelity of outputs with SD 1.5 and SD 2.1 is lower than those generated with the more advanced SDXL~\cite{podell2023sdxl}.

\begin{figure*}[!t]
    \centering
    \includegraphics[width=\linewidth]{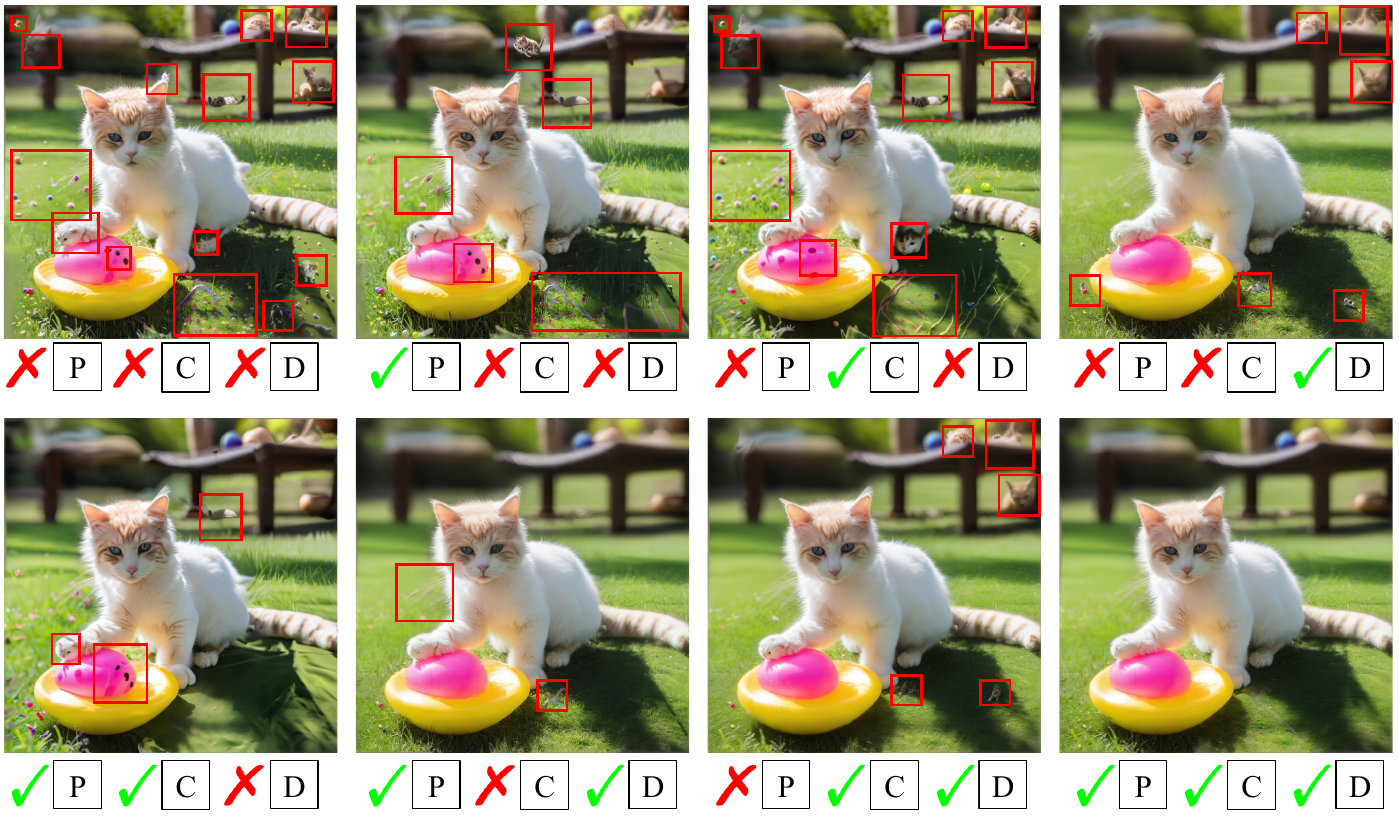}
    \caption{
    Ablations of Patch-content-aware prompts (\boxed{\text{P}}), ControlNet assisted generation (\boxed{\text{C}}), and Dilated sampling with window interaction (\boxed{\text{D}}). The ``{\color{red}\XSolidBrush}''/``{\color{green}\Checkmark}''  denotes removing/preserving the component. The artifacts are highlighted by a red box. The prompt of image is ``A cat is playing with furry toys on the lawn.''. Best viewed zoomed in.}
    \label{fig:acc_ablation}
\end{figure*}

\subsection{Ablation Study}
\label{ablaiton study}
This section begins with ablation studies on the three core modules introduced in this paper, followed by a discussion on the threshold settings for the binary mask in Eq.\,(\ref{eq:highly responsive regions}) and the patch-content-aware prompt threshold $c$ in Eq.\,(\ref{eq:patch-content-aware prompt}). All experiments use a resolution of $4096^2$ ($16 \times$). Since current quantitative metrics cannot intuitively reflect the extent of object repetition or local distortion, we provide visualizations to show how our core modules effectively prevent repetitive generation and local distortion.

\subsubsection{Ablations on Core Modules} 
\label{sec:ablation on core modules}
Fig.\,\ref{fig:acc_ablation} illustrates that removing any module reduces generation quality. 
Excluding patch-content-aware prompts leads to numerous small, repetitive object repetitions, emphasizing the role of patch-content-aware prompts in preventing repetitive generation.
When dilated sampling with window interaction is removed, small objects in the image appear unrelated to the image, demonstrating that dilated sampling with window interaction enhances semantic consistency and minimizes repetition.
Removing dilated sampling with window interaction leads to small object occurrences in parts of the big object, \emph{e.g.}, a small cat head on the claw of the big cat.
This indicates that dilated sampling with window interaction helps maintain the overall structure of the image by providing accurate global structure information.
Moreover, without ControlNet-assisted generation, the image exhibits local distortion, indicating that ControlNet helps establish more accurate local structures.
When all modules are removed, the image shows the most repetitive objects; however, using all modules together effectively prevents both repetitions and local distortion. This demonstrates that these modules function collectively to minimize artifacts.

\begin{table}[!htbp]
    \caption{Statistics of cross-attention maps $\mathcal{M}$ using prompt $y$ = ``Astronaut on mars during sunset.'' as an example. Each word $\{y_j\}^6_{j=1}$ has a cross-attention map ${\{\mathcal{M}}_{:,j}\}^6_{j=1}$.}
    \resizebox{\linewidth}{!}{
    \centering
    \setlength\tabcolsep{3pt}
    \begin{tabular}{@{}ccccccc@{}}
        \toprule
        \multirow{2}{*}{Statistics}  & ``Astronaut'' & ``on''    & ``mars'' & ``during'' & ``sunset'' & ``.''   \\
                               &   $(j=1)$     &  $(j=2)$  &  $(j=3)$ &  $(j=4)$   &  $(j=5)$   & $(j=6)$ \\
         \midrule
         $ \text{Min}({\mathcal{M}}_{:,j})$ & 0.1274 & 0.0597 & 0.2039 & 0.0457 & 0.0921 & 0.0335 \\
         $ \text{Mean}({\mathcal{M}}_{:,j})$ & 0.1499 & 0.0676 & 0.2533 & 0.0521 & 0.1189 & 0.0386 \\
         $\text{Max}({\mathcal{M}}_{:,j})$ & 0.2096 & 0.0779 & 0.2979 & 0.0585 & 0.1499 & 0.0419 \\
         \bottomrule
    \end{tabular}
    }
    \label{tab:ablation for highly responsive threshold}
    \vspace{-1em}
\end{table}

\subsubsection{Ablations on Hyper-Parameters}

\begin{figure*}[!t]
    \centering
    \includegraphics[width=\linewidth]{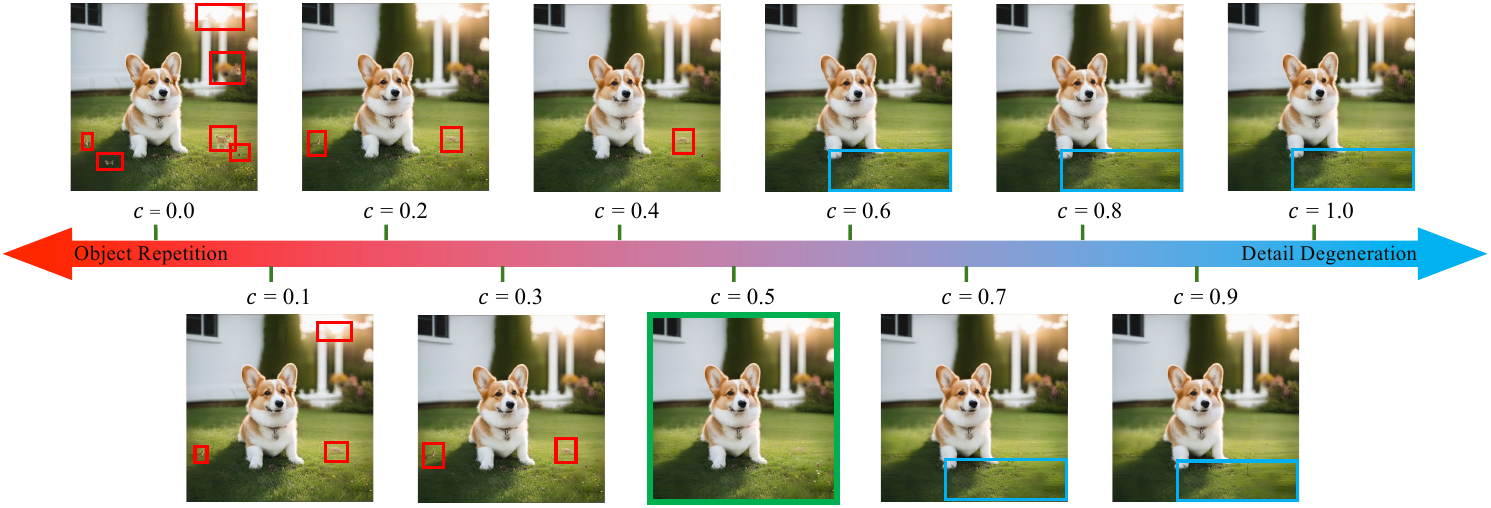}
    \caption{Visual results of different threshold $c$, prompted by ``A cute corgi on the lawn.'' The repetitive objects are highlighted with a red box and the detail degradation is stressed with a blue box. The best trade-off between object repetition and detail degradation is highlighted in the green box. Best viewed zoomed in.}
    \label{fig:different_c}
\end{figure*}

Table\,\ref{tab:ablation for highly responsive threshold} illustrates a significant variation in the range of different cross-attention maps \(\mathcal{M}_j\). 
Two potential scenarios arise when a fixed threshold is applied to these maps. 
In the first case, if the threshold is set too high, some words may lack highly responsive regions in their corresponding attention maps, leading to their exclusion from the patch-content-aware prompt. 
In the second case, if the threshold is set too low, the entire attention map may consist of highly responsive regions, resulting in those words being included in the patch-content-aware prompt all the time. 
By taking into account the average $\overline{\mathcal{M}}_{:,j}$, we can ensure that each word is associated with appropriate highly responsive regions, as shown in Fig.\,\ref{fig:attention_map}(b).

Referencing Eq.\,(\ref{eq:patch-content-aware prompt}), the parameter $c$ dictates whether the percentage of a highly responsive region for a word $y_j$ exceeds the threshold necessary for incorporation into the prompts of patch $z_t^i$.
When $c$ is set to a very small value, more words are incorporated into the patch prompt, potentially resulting in object repetition.
On the contrary, a significantly large value for $c$ simplifies the patch prompt, potentially leading to a loss of detail.
The demonstration of our analysis is depicted in Fig.\,\ref{fig:different_c}.
It is essential to recognize that this hyper-parameter is tailored to individual users and can be adjusted to fit various application scenarios.

\begin{figure*}[!t]
    \centering
    \includegraphics[width=\linewidth]{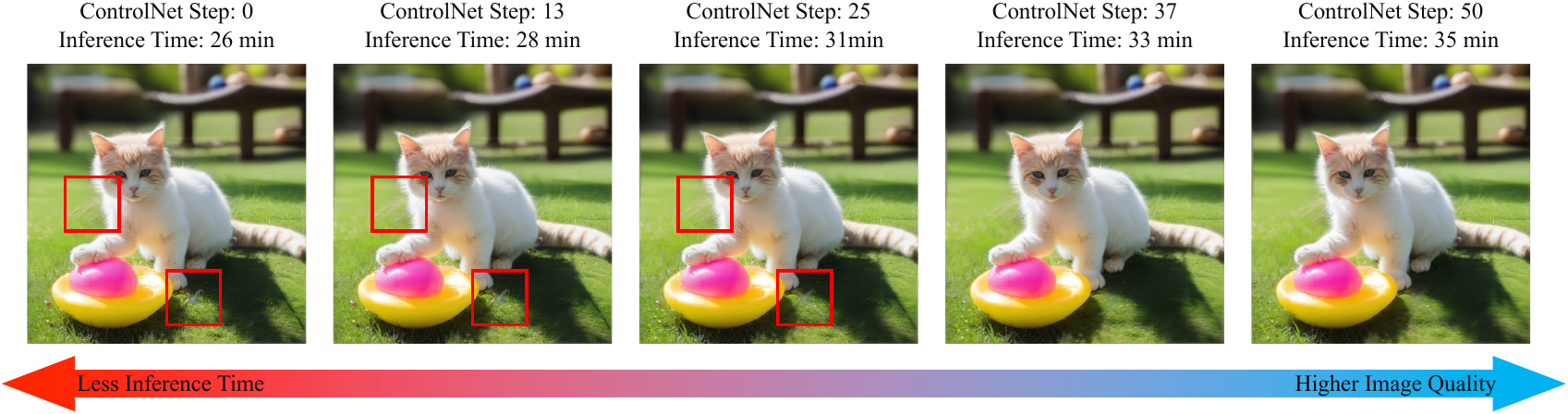}
    \caption{Ablation study on the number of ControlNet steps. Local distortions are highlighted with red boxes.
    The inference time is measured on one NVIDIA RTX 3090 GPU.
     The prompt used for the image is ``A cat is playing with furry toys on the lawn.'' Best viewed by zooming in.   \label{fig:ablation_controlnet_step}}
\end{figure*}

\subsubsection{Ablations on ControlNet Integration}
In Sec.\,\ref{sec:controlnet}, we integrate ControlNet into our framework to enhance the fidelity of local structures during higher-resolution extrapolation.
While this integration introduces additional computational overhead, it suppresses local artifacts in the results.
To rigorously evaluate the trade-off between image quality and inference efficiency, we perform an ablation study by varying the number of ControlNet steps. 
%
Specifically, we alternate between AccDiffusion v2 without ControlNet and AccDiffusion v2 with ControlNet during denoising, allowing us to flexibly adjust the frequency of ControlNet invocation.
For example, when the ControlNet steps are set to 25, it is applied once every two steps. 
As shown in Fig.\,\ref{fig:ablation_controlnet_step}, increasing the number of ControlNet steps yields notable improvements in image quality, primarily through the suppression of local artifacts. 
However, this enhancement comes at the expense of increased inference time.
Accordingly, the number of ControlNet steps can be adjusted to balance perceptual quality and inference efficiency based on application requirements.

The additional computational cost in AccDiffusion v2 arises from the inclusion of ControlNet’s condition encoder. 
To mitigate this, one potential solution is the adoption of lightweight alternatives such as ControlNeXt~\cite{peng2024controlnext}, which maintains comparable guidance performance with reduced latency. 
Moreover, on high-end GPUs equipped with enhanced parallelism (\emph{e.g.}, NVIDIA H100), the relative overhead introduced by ControlNet is significantly alleviated.
For instance, in the $16\times$ extrapolation setting, AccDiffusion v2 completes generation in 11 minutes on an H100 GPU, compared to 9 minutes for AccDiffusion.

\begin{figure}[!h]
    \centering
    \includegraphics[width=\linewidth]{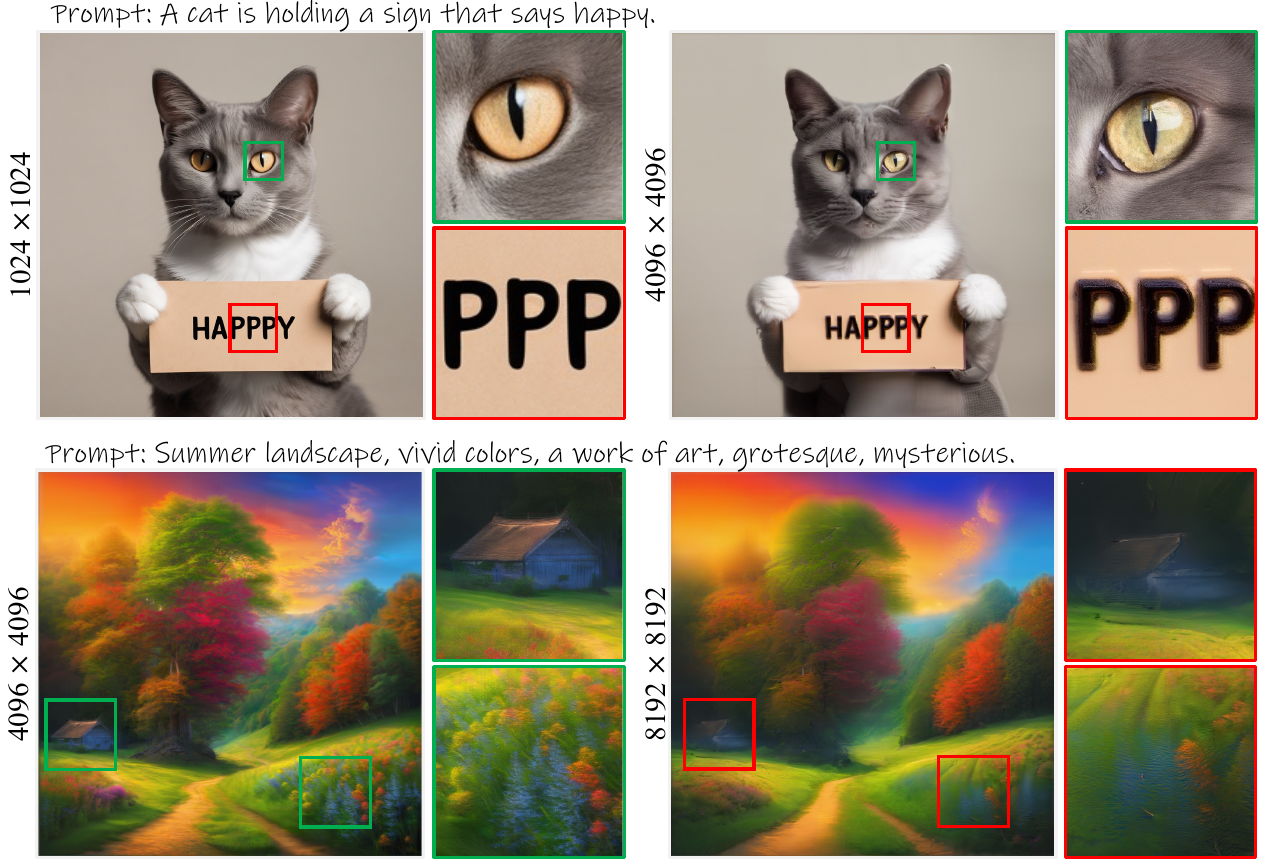}
    \caption{
Failure cases of AccDiffusion v2. The bad details are highlighted in a red box, while the good details are highlighted in a green box. Best viewed by zooming in.}
    \label{fig:failure cases}
\end{figure}

\section{Limitations and Future work}
While providing valuable insights, AccDiffusion v2 is not without its shortcomings: Firstly, the inference latency of AccDiffusion v2 is high as shown in Table\,\ref{tab:quantitative comparison}, akin to other patch-wise extrapolation methods~\cite{lin2024accdiffusion, du2023demofusion}, due to inefficient progressive upscaling and overlapped patch-wise denoising. Additionally, the use of ControlNet to suppress local distortion further adds to this delay.
Secondly, the fidelity of extrapolation results heavily relies on the pre-trained diffusion model, given that AccDiffusion v2 is training-free. 
Consequently, stronger diffusion models lead to improved AccDiffusion v2 performance and vice versa.
Thirdly, as depicted in Fig.\,\ref{fig:failure cases}, AccDiffusion v2 excels in generating intricate details like the cat's eye but may introduce irrelevant elements such as superfluous ``PPP'' details.
Lastly, both AccDiffusion and patch-wise methods should allow infinite extrapolation. However, when the resolution exceeds 8K (64×), AccDiffusion v2, along with existing techniques~\cite{du2023demofusion, lin2024accdiffusion}, encounters detail degradation.

To enhance efficiency, forthcoming research could explore non-overlapping patch-based denoising techniques and use a lightweight condition encoder to alleviate inference latency. 
AccDiffusion v2 offers valuable insights, decoupling the image-content-aware prompt into patch-content-aware prompts to address object repetition caused by inaccurate prompts. 
Future works have an opportunity to use vision large language models~\cite{liu2023llava} or image caption models~\cite{li2023blip} to refine prompts for distinct patches.
Furthermore, AccDiffusion v2 highlights the benefits of incorporating extra controls like ControlNet~\cite{zhang2023controlnet} to generate coherent high-resolution image structures. 
Future studies could delve into employing controllable generation methods for high-resolution image extrapolation.

\section{Conclusion}
This paper proposes AccDiffusion v2, a plug-and-play module, that enables higher-resolution diffusion extrapolation without repetitive generation or local distortion. 
To improve patch-wise denoising accuracy, AccDiffusion v2 introduces patch-content-aware prompts, effectively addressing the issue of repetitive generation from the root.
Additionally, to mitigate local distortion, AccDiffusion v2 integrates more precise local structural information through ControlNet during the higher-resolution diffusion extrapolation.
Moreover, we propose dilated sampling with window interaction to improve global consistency while generating high-resolution images.
Comprehensive experiments demonstrate that AccDiffusion v2 achieves state-of-the-art performance, successfully generating higher-resolution images without object repetition or local distortions.

\section*{Acknowledgments}
This work was supported by  the National Science Fund for Distinguished Young Scholars (No.62025603), the National Natural Science Foundation of China (No. U21B2037, No. U22B2051, No. U23A20383, No. 62176222, No. 62176223, No. 62176226, No. 62072386, No. 62072387, No. 62072389, No. 62002305 and No. 62272401), and the Natural Science Foundation of Fujian Province of China (No. 2021J06003, No.2022J06001).



\ifCLASSOPTIONcaptionsoff
  \newpage
\fi




\bibliographystyle{IEEEtran}
\bibliography{main}

\end{document}